\documentclass[lettersize,journal]{IEEEtran}
\usepackage{amsmath,amsfonts}
\usepackage{algorithmic}
\usepackage{algorithm}
\usepackage{array}
\usepackage[caption=false,font=normalsize,labelfont=sf,textfont=sf]{subfig}
\usepackage{textcomp}
\usepackage{stfloats}
\usepackage{url}
\usepackage{verbatim}
\usepackage{graphicx}
\usepackage{cite}
\usepackage{color}
\usepackage{url}
\usepackage{hyperref}

\usepackage{pdfpages}

\begin{document}

\title{Dendrite Net: A White-Box Module for Classification, Regression, and System Identification}

\author{Gang~Liu,~\IEEEmembership{}
	    Jing~Wang

\IEEEcompsocitemizethanks{\IEEEcompsocthanksitem G. Liu is with Institute of Robotics and Intelligent Systems, iHarbour Academy of Frontier Equipment, Xi’an Jiaotong University, Shaanxi, 710049, China (\textit{G. Liu presented DD.e-mail: gangliu.6677@gmail.com}).
	\IEEEcompsocthanksitem J. Wang is with Institute of Robotics and Intelligent Systems,, iHarbour Academy of Frontier Equipment, Xi’an Jiaotong University, Shaanxi, 710049, China (\textit{J. Wang offered advice.e-mail: wangpele@gmail.com)}.}
\thanks{\textit{\textbf{Gang Liu presented Dendrite Net, Relation Spectrum, and Gang neuron. His current research interests include machine learning, deep learning, computer vision, natural language processing, brain-computer interface, neuromorphic computing, and neurorehabilitation.
	Welcome to cooperate.}}}}

\markboth{Published by IEEE Transactions on Cybernetics}%
{Shell \MakeLowercase{\textit{et al.}}: A Sample Article Using IEEEtran.cls for IEEE Journals}


\maketitle

\begin{abstract}
	
The simulation of biological dendrite computations is vital for the development of artificial intelligence (AI). This paper presents a basic machine learning algorithm, named Dendrite Net or DD, just like Support Vector Machine (SVM) or Multilayer Perceptron (MLP). DD's main concept is that {\color{magenta} the algorithm can recognize this class after learning, if the output's logical expression contains the corresponding class's logical relationship among inputs} ({\color{magenta}\textit{and$\backslash$or$\backslash$not}}). Experiments and main results:  DD, a white-box machine learning algorithm,	showed excellent system identification performance for the  black-box system. Secondly, it was verified by nine real-world applications that DD brought better generalization capability relative to MLP architecture that imitated neurons' cell body (Cell body Net) for regression. Thirdly, by MNIST and FASHION-MNIST datasets, it was verified that DD showed higher testing accuracy under greater training loss than Cell body Net for classification. The number of modules can effectively adjust DD's logical expression capacity, which avoids over-fitting and makes it easy to get a model with outstanding generalization capability. Finally, repeated experiments in \textit{MATLAB} and \textit{PyTorch} (\textit{Python}) demonstrated that DD was faster than Cell body Net both in epoch and forward-propagation. The main contribution of this paper is the basic machine learning algorithm (DD) with a white-box attribute, controllable precision for better generalization capability, and lower computational complexity. Not only can DD be used for generalized engineering, but  DD has vast development potential as a module for deep learning. DD code is available at \textit{GitHub:\href{https://github.com/liugang1234567/Gang-neuron}{Gang neuron}}.

\end{abstract}


\section{Introduction}
\IEEEPARstart{T}{he} simulation of biological neuron computations has long been a question of great interest in a wide range of fields. In 70 years ago, researchers thought that biological dendrites did not perform logic operations \cite{mcculloch1943logical}. Therefore, McCulloch and Pitts proposed a simple neuron model ``$ f(wx+b) $". Today, it was discovered that the previous neuron model is only a cell body model. Nowadays, biological dendrites in brains have been proven to have \textit{and$\backslash$or$\backslash$xor} logic operations \cite{gidon2020dendritic,london2005dendritic,mel1994information,shepherd1987logic}. The simulation of dendrite computations is to realize the multiple logical operations in essence. The multiple-valued logic network (MVL) proposed by Zheng Tang et al. in 1988 is a classic algorithm for multiple logical operations\cite{tang1998learning}. MVL consists of three basic operations denoted by ``$ + $" (sum), ``$ \cdot $" (multiplication), and ``$ f $" (piecewise linear operation). In 2019, Yuki Todo et al. optimized MVL using a sigmoidal thresholding nonlinear operation to simulate biological dendrites and proposed neurons with multiplicative interactions of nonlinear synapses\cite{todo2019neurons}. In 2020, Jian Sun  et al. extended MVL to multiobjective optimization algorithm from single-objective optimization \cite{sun2018bi}.
\begin{figure}[!t]
	\centering
	\includegraphics[width=\columnwidth]{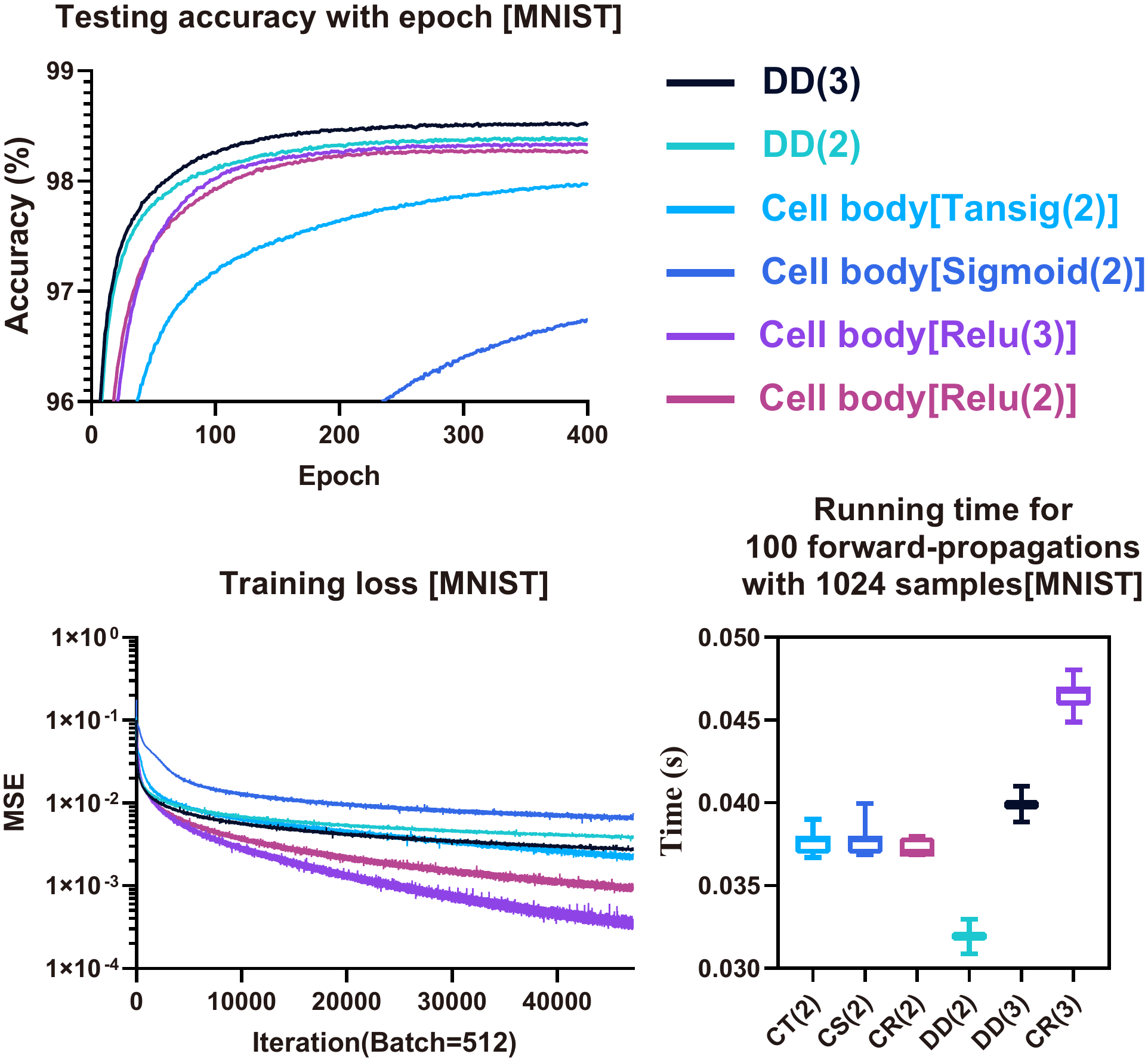}
	\caption{Testing accuracy, training loss, and computational complexity on MNIST with Cell body Nets and DDs. (1) Interestingly, DD shows higher testing accuracy under greater training loss. (2) DD shows the controllable training loss related to the number of DD modules, compared to the sustained decline of cell body’s training loss with iteration. (3) As the number of DD modules increases, training loss decreases, and testing accuracy increases. (4) DD shows lower computational complexity in the same number of parameters. Similar phenomenons are presented in Fig. 12 and Fig.14. [ We stress the basic module's essential properties instead of accuracy because all modules can be inserted into larger architecture for outstanding performance. These properties are base.]}
	\label{fig1}
\end{figure} 
Additionally, in 2019, Shangce Gao et al. proposed a dendritic neuron model (DNM) with synaptic layer, dendrite layer, membrane layer, and soma layer and solved the parameters of DNM by biogeography-based optimization, particle swarm optimization, genetic algorithm, ant colony optimization, evolutionary strategy, and population-based incremental learning \cite{gao2018dendritic}. However, both MVL-based algorithms and DNM contain a function (``$ f $") and are too complicated. Furthermore, the literature\cite{durbin1989product} showed a product unit with multiplication and addition. Nevertheless, its expressive ability is insufficient, it can only express the first-order input, and it is very complicated for the computer to realize it.

Based on the characteristic of biological dendrites, the simulation of dendrite computations should be as straightforward as possible. In order to combine with a cell body in the existing neural networks in the future, the optimization algorithm is better if it is error-backpropagation.  In addition, perhaps we should implement dendrites' functions (\textit{and$\backslash$or$\backslash$xor}) in the simplest and most conducive form for computer operation, instead of pursuing the physical form of dendrites. After all, the computing form of the computer and the brain are different; however, the functions can be the same. Too much pursuit of the shape of dendrites may cause a lot of extra calculations for the computer.

\begin{figure*}[!t]
	\centering
	\includegraphics[width=1.7\columnwidth]{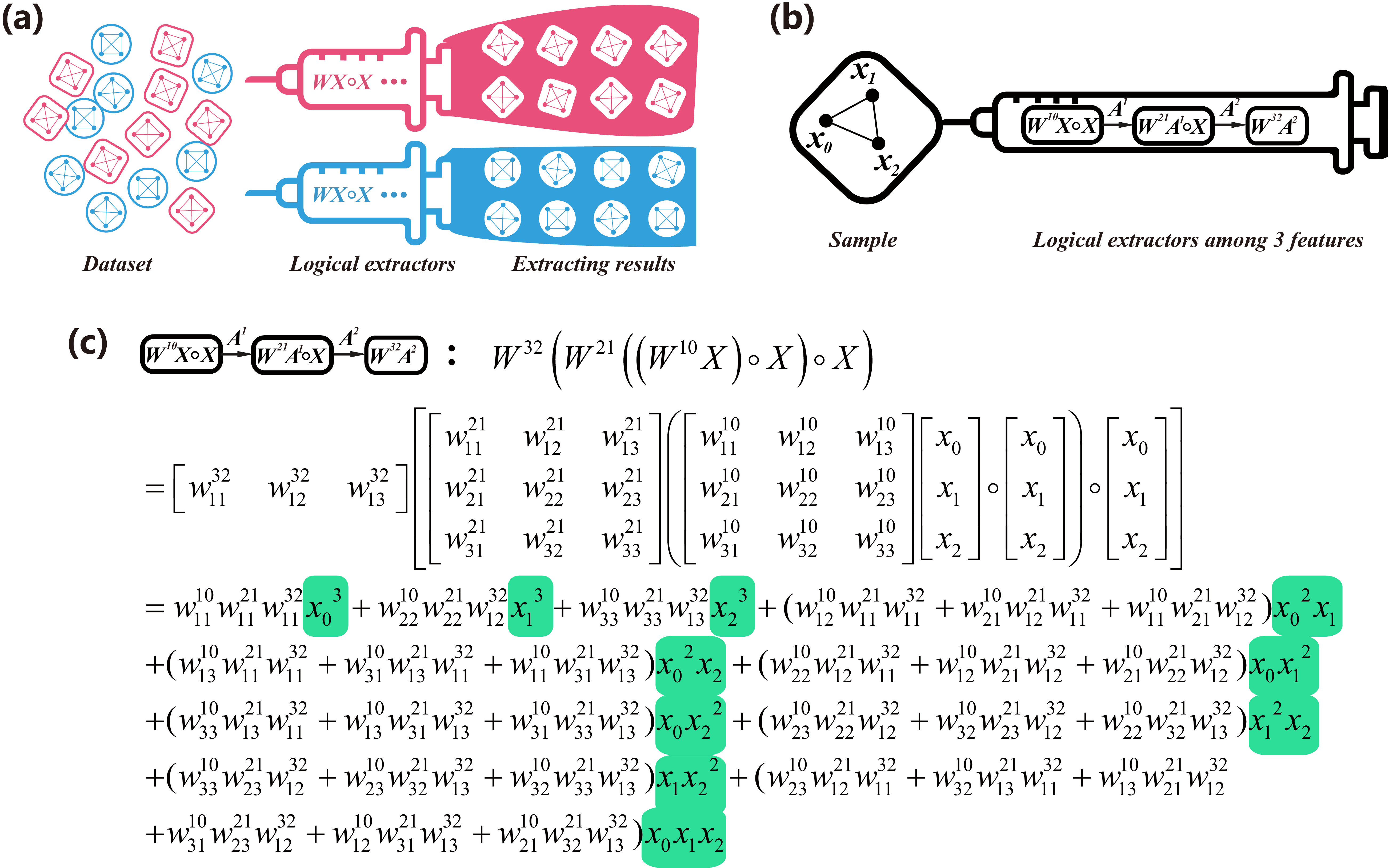}
	\caption{Interpretation of DD. (a) DD aims to design the output's logical expression of the corresponding class (logical extractor). DD’s logical extractors look like some graduated pipettes. When an unknown sample appears, the logical extractor that extracts more information `` grabs" this sample. (b) DD achieves the controllable precision of the logical extractor by the number of DD modules. The figure shows an example using 3 modules (The last one is a linear module). (c) The expanded form of the logical extractor in (b). DD contains all logical relationships information under a certain degree. $ x_{0} $ can be set as $ 1 $.  ``and'': multiplication (e.g.,$ x_{1}x_{2} $ ); ”or”: addition (e.g., $ x_{1}+x_{2} $ );”not”: minus (e.g., $ -x_{1} $ or $ -x_{2} $ ).}
	\label{fig2}
\end{figure*}

Additionally, the definition of a many-valued logic (also multi- or multiple-valued logic) is a propositional calculus in which there are more than two truth values\cite{bolc2013many,rosser1957many}. Therefore, we can redefine the expression of multiple logical operations instead of using the MVL in 1988.  ``\textit{xor}'' can be expressed by basic logical element (\textit{and$\backslash$or$\backslash$not}). Thus, we can define multiple logical operations which only contain basic logical elements (\textit{and$\backslash$or$\backslash$not}). Besides, a study showed that the integration of simultaneous excitatory postsynaptic potentials (EPSP) and inhibitory postsynaptic potentials (IPSP) could be described well in multiplicative form \cite{hao2009arithmetic}. Therefore, this paper presented DD that only contains matrix multiplication and Hadamard product. There is no doubt that the logical relationship (\textit{and$\backslash$or$\backslash$not}) among features determines the sample's class \cite{quinlan1990learning}. DD is to extract the amount of logical relationship information. If one class's output expression contains its logical relationship information among features, this output expression can be regarded as the corresponding class's logical extractor according to many-valued logic theory \cite{rosser1957many,bolc2013many}. In this way, each class has a logical extractor. These extractors, like a graduated pipette, can extract logical information from the data. When an unknown sample appears, the class whose logical extractor extracts more information can ``grab" this sample (see Fig. \ref{fig2}). 

The main contribution of this paper is the basic machine learning algorithm (DD) with the white-box attribute, controllable precision for better generalization capability, and lower computational complexity. Additionally, DD is first proposed in this paper. As a basis for new studies in the future \cite{Liu2020}, this paper focuses on DD's characteristics in terms of a basic ML algorithm. The remainder of this paper is divided as follows: Related work, DD, System identification, Regression, Classification, Computational complexity, Additional discussion, and Conclusion.

\section{Related work}

\subsection{Basic ML Algorithms}
So far, some basic ML algorithms have been proposed. \textbf{However, there is no DD.} This section briefly reviews these basic algorithms. (\textit{See supplementary materials.})

The review contains Least Squares Regression, Logistic Regression, Linear discriminant analysis (LDA), Decision Trees, Naive Bayes Classification, Naive Bayes Classification, Support Vector Machines, Ensemble Methods, Clustering Algorithms, Random forests or random decision forests, and Artificial Neural Networks (ANN).

It is worth emphasizing that the above algorithms have their own advantages in different aspects; however, there is no DD. The benefits of DD, such as white-box attribute, controllable precision for better generalization capability, and lower computational complexity, might pose new changes in many fields in the future.

\subsection{Easily-confused Work}

\textbf{Self-attention mechanism \cite{vaswani2017attention}:} From a particular perspective, DD can be regarded as a self-attention module. However, DD is simpler and ``rude''. All parameters are in one weight matrix and rely on the self-adaption of the model. The more modules are connected in series, and the more detailed features are extracted. The simpler architecture makes DD easier to use and becomes a white-box algorithm.

\textbf{ANN with polynomial activation function \cite{ma2005constructive,jiang2016potential}:} DD is \textbf{not} an ANN with a polynomial activation function. A polynomial activation function is a non-linear function with a definite form, and the inputs are as independent variables of the function. However, DD is with Hadamard product between the current inputs and previous inputs. DD and model in \cite{ma2005constructive,jiang2016potential} are entirely different. It's just that the diagram drawn is somewhat similar. The diagram represents a completely different meaning. \cite{ma2005constructive,jiang2016potential} also showed an ANN with a polynomial activation function. Of course, DD can be regarded as a function whose form is indeterminate. Nevertheless, we think it is unreasonable to regard it as an activation function. We should not force it into the previous concept. There is no activation function in DD. Hadamard product is used to establish logical relationships among inputs.

\textbf{Polynomial regression (PR) \cite{schielzeth2010simple,poggio1975optimal}:} From the expanded form, DD looks like a polynomial regression. Nevertheless, there are four obvious differences. (1) Traditional multiple regression cannot be applied to classification problems. (2) Traditional multiple regression is converted into linear regression through linear processing. Then, the least square method or error backpropagation is used to solve the preset parameters, which is different from DD. (3) As the order increases, the computational complexity of PR models increases exponentially. However, for each additional order, DD only needs to add one module, no matter how many modules DD has currently. (4) PR has only one output value, but DD can have multiple outputs.

A basic algorithm is an information processing method essentially. Therefore, all algorithms are somewhat similar from some particular perspective. DD contains all logical relationships information under a certain degree. It is very simple \textbf{{\color{magenta}[only one line of code: $ X=W@X*X $ (\textit{Python})]}} and is suit for problems with a large number of features like image classification. Because of DD’s straightforward architecture, one can imagine its potential for development and application in engineering fields. 

\section{DD}

Uppercase letters denote a matrix, and lowercase letters denote an element in the following formulas.

\subsection{Architecture}

\begin{figure}[!t]
	\centering
	\includegraphics[width=0.7\columnwidth]{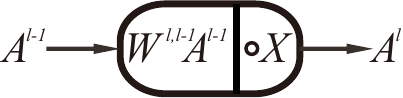}
	\caption{DD module. }
	\label{fig3}
\end{figure}

DD consists of DD modules and linear modules. The DD module is straightforward (see Fig. \ref{fig3}) and is expressed as follows.

\begin{equation} 
	A^{l}=W^{l,l-1}A^{l-1} \circ X
	\label{eq.1}
\end{equation}
where $ A^{l-1} $ and $ A^{l} $ are the inputs and outputs of the module. $ X $ denotes the inputs of DD. $ W^{l,l-1} $ is the weight matrix from the  $ (l-1)$-th module to the $ l$-th module.” $ \circ $ denotes Hadamard product. Hadamard product is a binary operation that takes two matrices of the same dimensions and produces another matrix of the same dimension as the operands, where each element $ i $, $ j $ is the product of elements $ i $, $ j $ of the original two matrices. 

\begin{figure*}[!t]
	\centering
	\includegraphics[width=1.5\columnwidth]{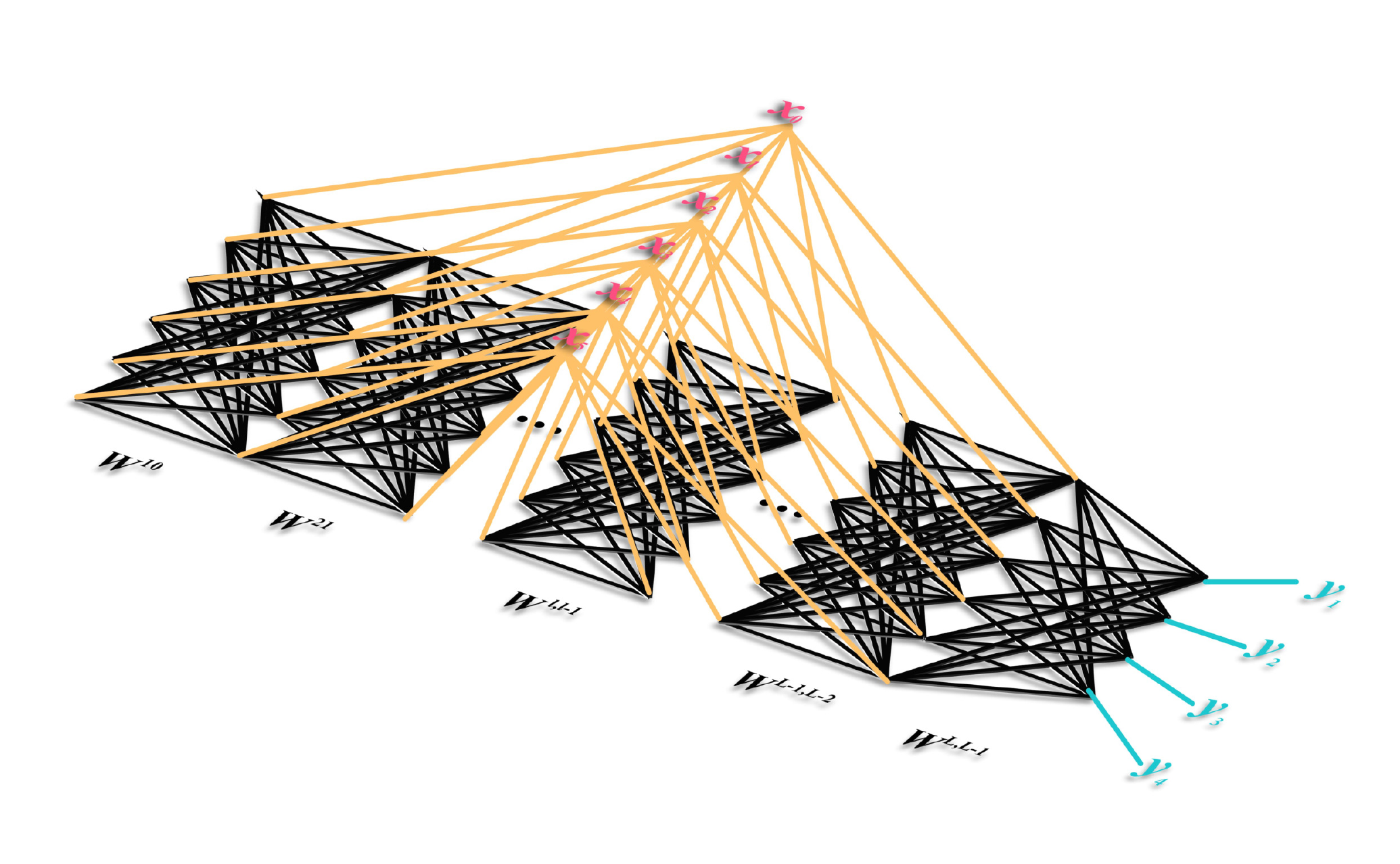}
	\caption{The overall architecture of DD with 6 inputs and 4 outputs. The figure is the example visualization of Eq \ref{eq.2}.}
	\label{fig4}
\end{figure*}

The overall architecture of DD is shown in Fig. \ref{fig4}. The architecture can be represented according to the following formula.

\begin{equation} 
	\begin{split}
		\label{eq.2}
		Y=W^{L,L-1}[\cdots W^{l,l-1}(\cdots W^{21}(W^{10}X \circ X) \circ X \cdots) \circ X\\
		\cdots], L\in N+
	\end{split}
\end{equation}
where $ X $ and $ Y $ denote the input space and the output space. $ W^{l,l-1} $ is the weight matrix from the  $ (l-1)$-th module to the $ l$-th module. The last module is linear.   $ L $ expresses the number of modules. The calculation of DD only contains matrix multiplication and Hadamard product. It is well known that the computational complexity of Hadamard product is significantly lower than non-linear functions.

\subsection{Learning Rule}

\begin{figure*}[!t]
	\centering
	\includegraphics[width=2\columnwidth]{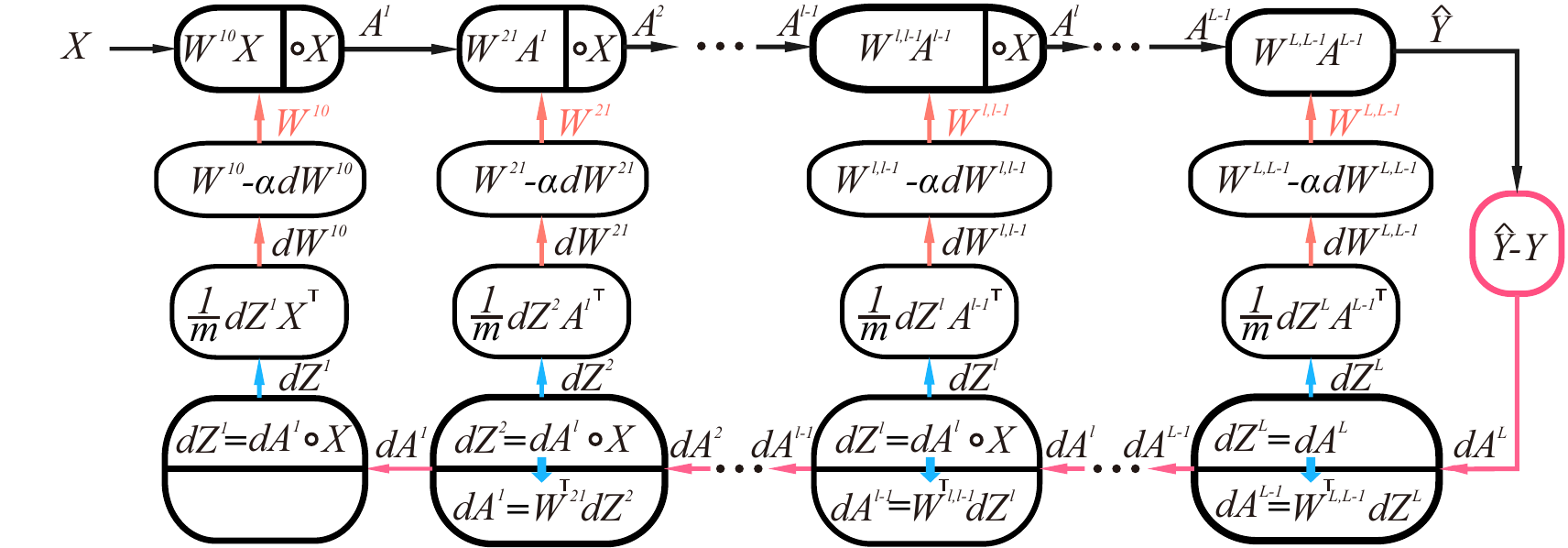}
	\caption{Graphical illustration of learning rule.}
	\label{fig5}
\end{figure*}

The graphical illustration of learning rule is shown in Fig. \ref{fig5}.  As an example, we use one-half of the mean squared error (MSE) as the loss function. We describe an error back-propagation-based learning rule for DD \cite{rumelhart1986learning}. The simplicity of the learning method makes it convenient for DD to be used in different areas. The following set of equations describes the simple gradient descent rule.

\textit{The forwardpropagation of DD module and linear module:}

\begin{equation} 
	\left\{
	\begin{aligned}
		A^{l}&=W^{l,l-1}A^{l-1} \circ X \\ 
		A^{L}&=W^{L,L-1}A^{L-1}
	\end{aligned}
	\right.
	\label{eq.3}
\end{equation}

\textit{The error-backpropagation of DD module and linear module:}

\begin{equation} 
	dA^{L}=\widehat{Y}-Y
	\label{eq.4}
\end{equation}
\begin{equation} 
	\left\{
	\begin{aligned}
		dZ^{L}&=dA^{L} \\ 
		dZ^{l}&=dA^{l} \circ X 
	\end{aligned}
	\right.
	\label{eq.5}
\end{equation}
\begin{equation} 
	dA^{l-1}=(W^{l,l-1})^{T}dZ^{l}
	\label{eq.6}
\end{equation}

\textit{The weight adjustment of DD:}

\begin{equation} 
	dW^{l,l-1}=\frac{1}{m}dZ^{l}(A^{l-1})^{T}
	\label{eq.7}
\end{equation}
\begin{equation} 
	W^{l,l-1(new)}=W^{l,l-1(old)}-\alpha dW^{l,l-1}
	\label{eq.8}
\end{equation}
where $ \widehat{Y} $ and $ Y $ are DD’s outputs and labels, respectively. $ m $ denotes the number of training samples in one batch. The learning rate $ \alpha $ can either be adapted with epochs or fixed to a small number based on heuristics.

\subsection{Theoretical proofs}

\textit{See supplementary materials.}

\subsection{Notes and Tips}

\textit{See supplementary materials.}

\section{System identification}

Unlike previous basic ML algorithms, DD is a white-box algorithm. The trained logical extractors of DD can be translated into the relation spectrum about inputs and outputs by formula simplification with software (e.g., \textit{MATLAB}).  Concretely, the optimized weights are assigned to the corresponding matrixes in Eq \ref{eq.2}. Then the relation spectrum was obtained through formula simplification of Eq \ref{eq.2} in software because DD only contains matrix multiplication and Hadamard product. The example and code can be found in GitHub. To illustrate the characteristics of DD intuitively, we show some simple and representative examples in this section.

\subsection{Identification comparison with Taylor's expansion}

As an example, we select three-modules DD with two inputs and one output. To illustrate the process of formula simplification in software, we show it with labour (\textit{See supplementary materials.}).
\begin{equation} 
	\resizebox{.9\hsize}{!}{$
		\begin{aligned}
			f(X)&= W^{32}(W^{21}((W^{10}X) \circ X)\circ X  ) \\ 
			&= w^{10}_{11}w^{21}_{11}w^{32}_{11}x_{0}^{3}+w^{10}_{22}w^{21}_{22}w^{32}_{12}x_{1}^{3}\\
			&\quad+(w^{10}_{12}w^{21}_{11}w^{32}_{11} + w^{10}_{21}w^{21}_{12}w^{32}_{11} + w^{10}_{11}w^{21}_{21}w^{32}_{12})x_{0}^{2}x_{1}\\
			&\quad+(w^{10}_{22}w^{21}_{12}w^{32}_{11} + w^{10}_{12}w^{21}_{21}w^{32}_{12} + w^{10}_{21}w^{21}_{22}w^{32}_{12})x_{0}x_{1}^{2}
		\end{aligned} 
		\label{eq.9}
		$}
\end{equation}
where $ x_{0} $ can be set as $ 1 $. Thus, the simplified DD consists of constant $ c $ , $ x_{1}^{3} $ , $ x_{1} $, and $ x_{1}^{2} $ items. The coefficients are in terms of DD's weights. For DD with more inputs and more modules, the result can be shown as a relation spectrum where the items and coefficients are the abscissa and ordinate, similar to Fourier spectrum.

\begin{figure}[!t]
	\centering
	\includegraphics[width=\columnwidth]{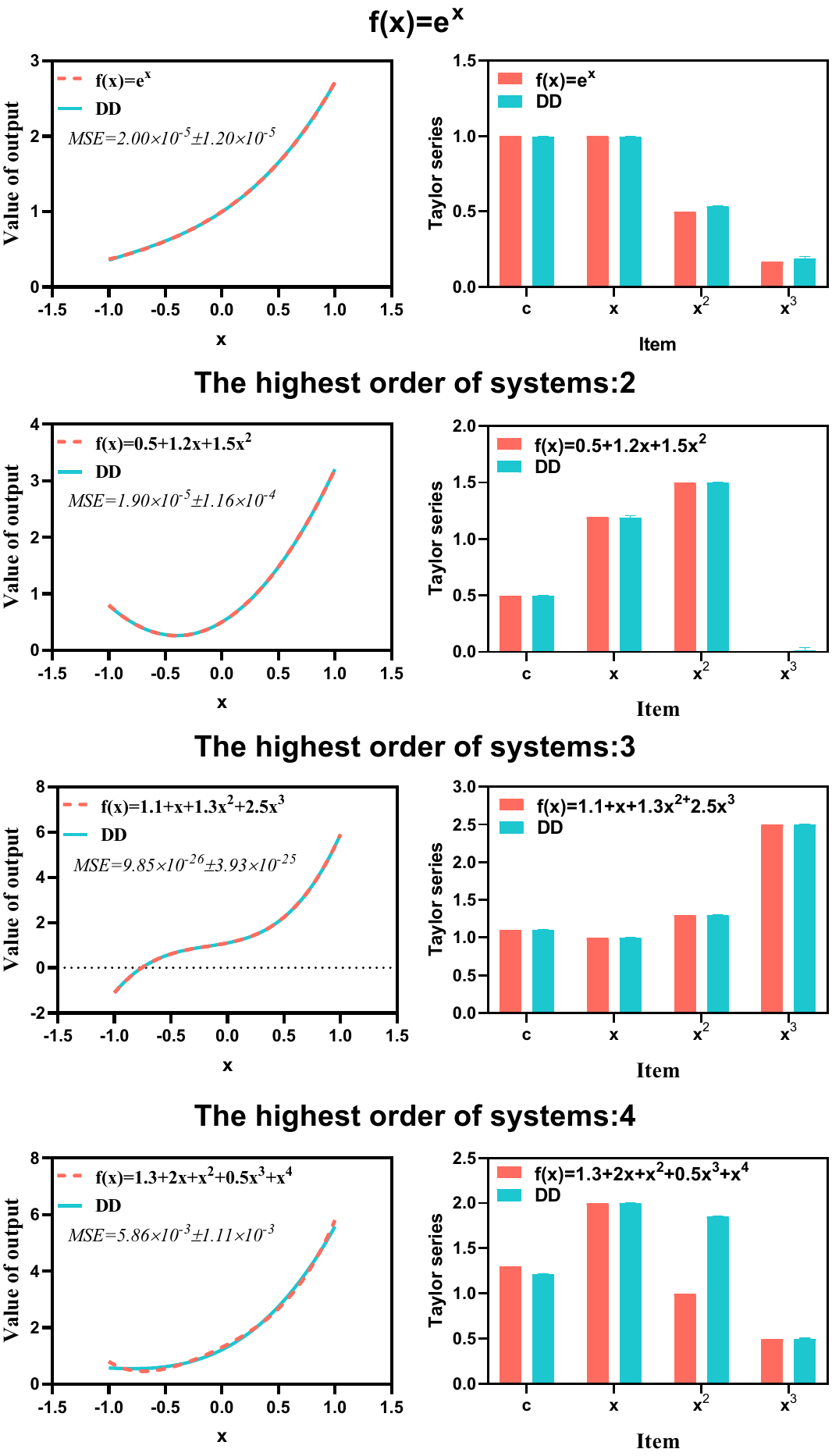}
	\caption{Identification comparison with Taylor's expansion. The left figures are the approximation to the function value. The right figures are the comparison between the real system and DD’s identifications. All DDs for $ 200 $ runs with different initial parameters. Taylor series: $ mean\pm SD $.}
	\label{fig7}
\end{figure}

To test whether the three-modules DD is similar to a three-order Taylor's expansion, we selected $ f(x)=e^{x} $ as an example. Furthermore, we compared DD's output with two-order, three-order, and four-order functions to explore the influence of DD modules' redundancy and inadequacy. For a more comprehensive comparison, we ran the algorithms $ 200  $ times using $ 200 $ different initial parameters for each DD, as shown in Fig. \ref{fig7}. These results indicate three attractive properties of DD.

(1) DD is similar to the Taylor expansion at the optimal combination points. (at $ 0 $ in Fig. \ref{fig7}). 

(2) DD can converge to the global optimum with high probability, as evidenced by the similar identifications in 200 runs.

(3) When the number of DD modules is inadequate, DD will search for the global optimum weights to approach the labels, as evidenced by the similar identification results in 200 runs for four-order functions.

\subsection{Identification for multiple-inputs system with noise}

As an example, we select three-modules DD with three inputs and one output. To illustrate the process of formula simplification in software, we show it with labour (\textit{See supplementary materials.}).

\begin{equation} 
	\resizebox{.9\hsize}{!}{$
		\begin{aligned}
			f(X)&= W^{32}(W^{21}((W^{10}X) \circ X)\circ X  ) \\ 
			&= w^{10}_{11}w^{21}_{11}w^{32}_{11}x_{0}^{3}+w^{10}_{22}w^{21}_{22}w^{32}_{12}x_{1}^{3}+w^{10}_{33}w^{21}_{33}w^{32}_{13}x_{2}^{3}\\
			&\quad+(w^{10}_{12}w^{21}_{11}w^{32}_{11} + w^{10}_{21}w^{21}_{12}w^{32}_{11} + w^{10}_{11}w^{21}_{21}w^{32}_{12})x_{0}^{2}x_{1}\\
			&\quad+(w^{10}_{13}w^{21}_{11}w^{32}_{11} + w^{10}_{31}w^{21}_{13}w^{32}_{11} + w^{10}_{11}w^{21}_{31}w^{32}_{13})x_{0}^{2}x_{2}\\
			&\quad+(w^{10}_{22}w^{21}_{12}w^{32}_{11} + w^{10}_{12}w^{21}_{21}w^{32}_{12} + w^{10}_{21}w^{21}_{22}w^{32}_{12})x_{0}x_{1}^{2}\\
			&\quad+(w^{10}_{33}w^{21}_{13}w^{32}_{11} + w^{10}_{13}w^{21}_{31}w^{32}_{13} + w^{10}_{31}w^{21}_{33}w^{32}_{13})x_{0}x_{2}^{2}\\
			&\quad+(w^{10}_{23}w^{21}_{22}w^{32}_{12} + w^{10}_{32}w^{21}_{23}w^{32}_{12} + w^{10}_{22}w^{21}_{32}w^{32}_{13})x_{1}^{2}x_{2}\\
			&\quad+(w^{10}_{33}w^{21}_{23}w^{32}_{12} + w^{10}_{23}w^{21}_{32}w^{32}_{13} + w^{10}_{32}w^{21}_{33}w^{32}_{13})x_{1}x_{2}^{2}\\
			&\quad+(w^{10}_{23}w^{21}_{12}w^{32}_{11} + w^{10}_{32}w^{21}_{13}w^{32}_{11} + w^{10}_{13}w^{21}_{21}w^{32}_{12}\\
			&\quad+(w^{10}_{31}w^{21}_{23}w^{32}_{12} + w^{10}_{12}w^{21}_{31}w^{32}_{13} + w^{10}_{21}w^{21}_{32}w^{32}_{13})x_{0}x_{1}x_{2}\\
		\end{aligned} 
		\label{eq.11}
		$}
\end{equation}
where $ x_{0} $ can be set as $ 1 $. Thus, the simplified DD consists of constant $ c $ , $ x_{1}^{3} $ , $ x_{2}^{3} $ , $ x_{1} $ , $ x_{2} $ , $ x_{1}^{2} $ , $ x_{2}^{2} $ , $ x_{1}^{2}x_{2} $ , $ x_{1}x_{2}^{2} $ , and $ x_{1}x_{2} $ items.

\begin{figure}[!t]
	\centering
	\includegraphics[width=\columnwidth]{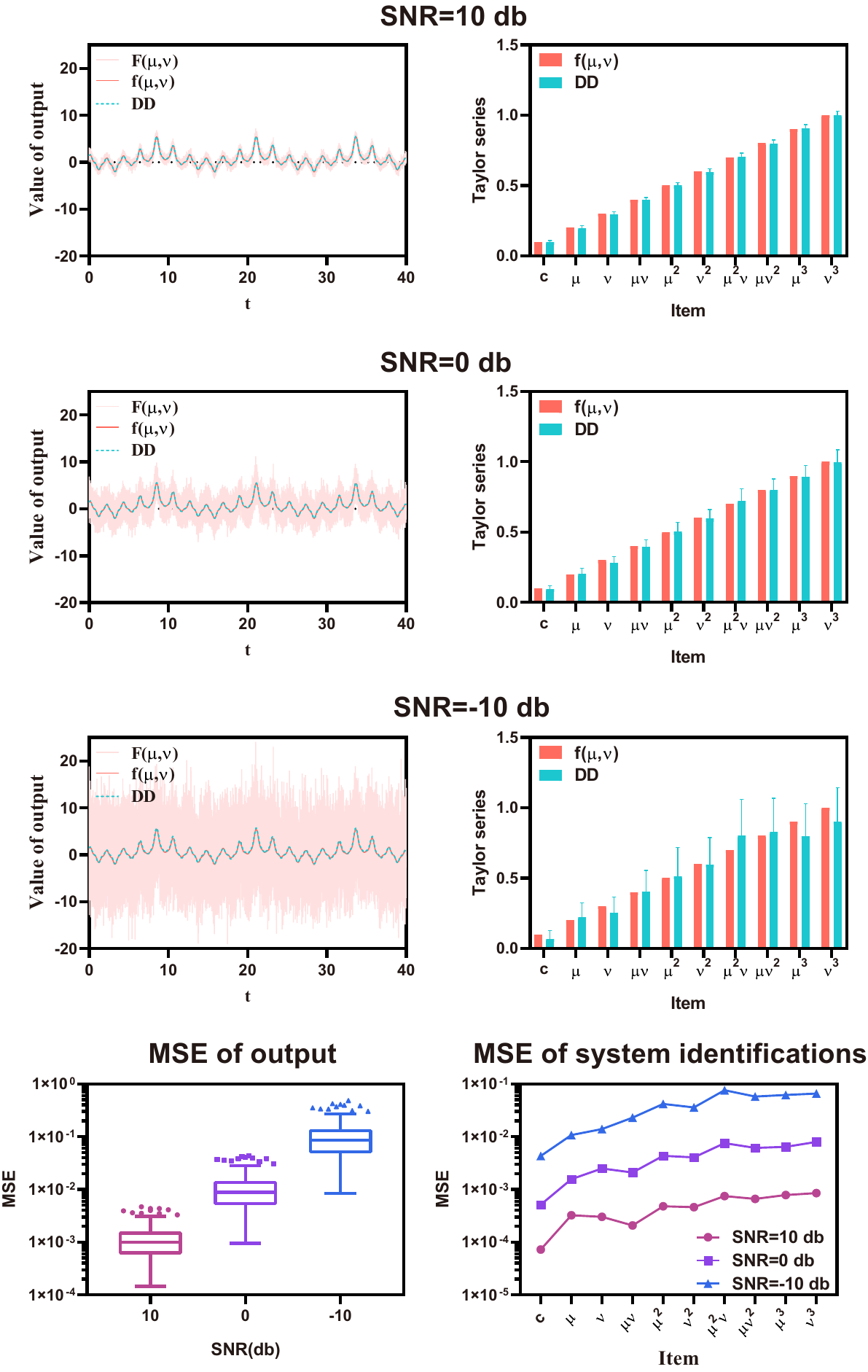}
	\caption{Identification for a multiple-inputs system with noise. The left figures are the approximation to the system output. The right figures are the comparison between the real system and DD’s identifications. All the DDs for 200 runs with different initial parameters in each case of SNR. Taylor series: $ mean\pm SD $.}
	\label{fig9}
\end{figure}

To assess DD's identification performance to a multiple-inputs system with noise, we constructed the three-order system.

\begin{equation} 
	\begin{split}
		\label{eq.12}
		f(u,v)=0.1+0.2u+0.3v+0.4uv+0.5u^2+0.6v^2\\
		+0.7u^2v+0.8uv^2+0.9u^2+v^3
	\end{split}
\end{equation}
Then, we added white Gaussian noise to $ f(u,v) $ to generate the output labels $ F(u,v) $ of DD.

\begin{equation} 
	F(u,v)=f(u,v)+N
	\label{eq.13}
\end{equation}
where $ N $ was the Gaussian noise. We explored identification performance with a signal-to-noise ratio (SNR) of -10, 0, and 10 $ db $, respectively, when the input is defined by:

\begin{equation} 
	\left\{
	\begin{aligned}
		u(t)=&sin\big( \frac{1}{2}t+60   \big) \\ 
		v(t)=&sin(3t+20)
	\end{aligned}
	\right.
	\label{eq.14}
\end{equation}
where we defined $ t \in [0,40] $. We ran DD 200 times using 200 different initial parameters for each condition, as shown in Fig. \ref{fig9}. It should be noted that we compared the DD model with the three-order system $ f(u,v) $ to be identified rather than the output labels $ F(u,v) $ . These results indicate two attractive properties of DD.

(1) DD shows excellent identification performance to a multiple-inputs system with noise even in the case of -10 SNR.

(2) As SNR decreased, MSE of system identification and output increased.  Interestingly, MSE of higher items increased greater than lower items with SNR decreased [Note that the ordinate is logarithmic coordinates in Fig. \ref{eq.9} (MSE)]. According to DD’s learning rule, the outline of output space is fitted by lower items, and later the higher items modify the details, which interprets this phenomenon.

\subsection{DD's approximation property}

For an unknown system, we should tune the numbers of DD modules to simulate the real system and then translate DD into a relation spectrum among input and output space. In order to investigate DD's approximation property, we considered the function fitting of the normalized Bessel function defined by:

\begin{equation} 
	f(x)=\frac{sin(x)}{x^{2}}-\frac{cos(x)}{x}
	\label{eq.10}
\end{equation}
where we defined  $ x \in [-10,0) \cup (0,10]$ , then $ x $ and $ f(x) $  were normalized to $ [-1,1] $ , respectively. 

\begin{figure}[!t]
	\centering
	\includegraphics[width=0.7\columnwidth]{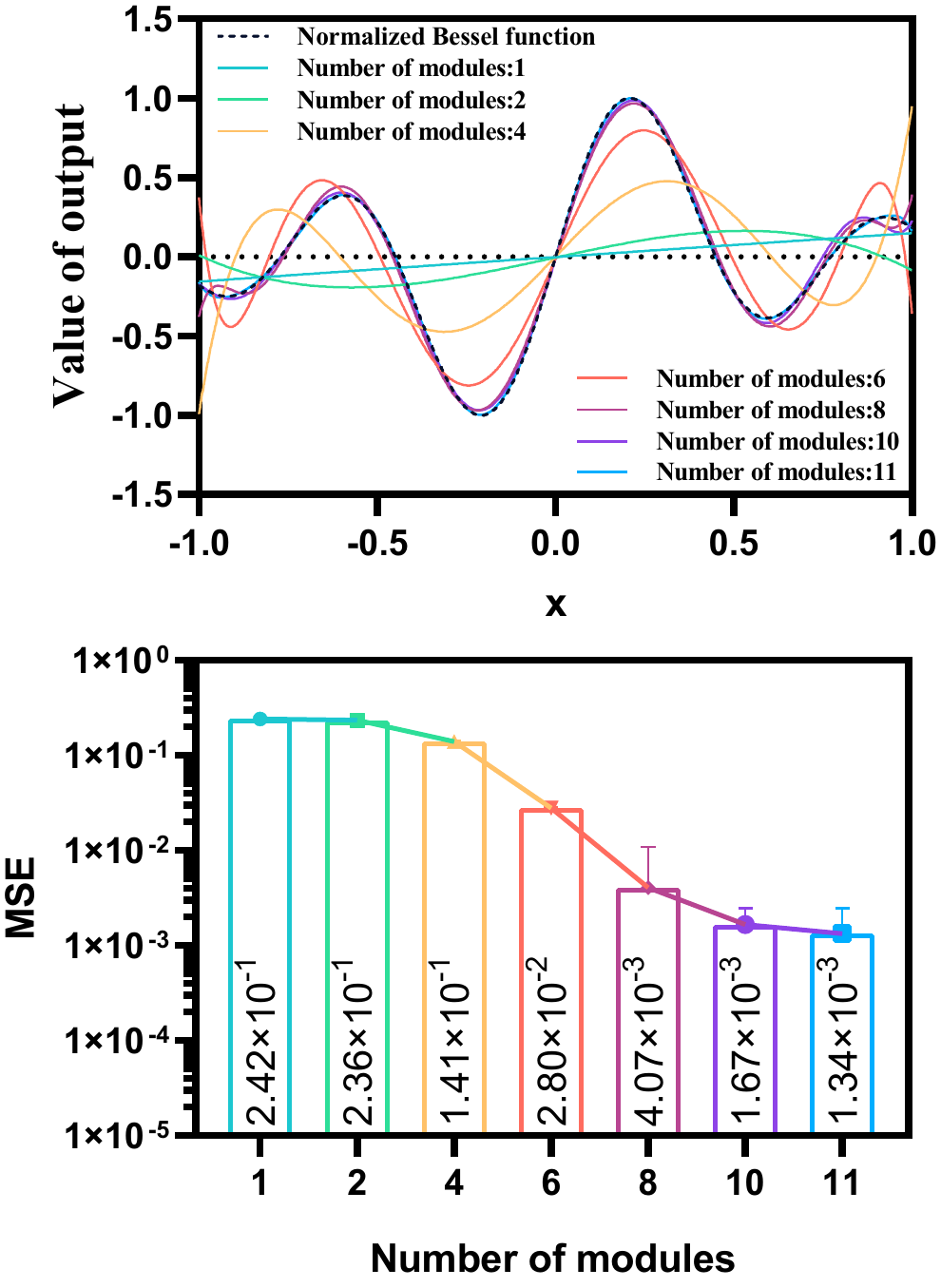}
	\caption{Approximation for normalized Bessel function. All DDs for 200 runs with different initial parameters. MSE: $ mean\pm SD $.}
	\label{fig8}
\end{figure}

We gradually increased DD modules to approach the normalized Bessel function. All DDs ran 200 times with different initial parameters (see Fig. \ref{fig8}). In this experiment, the accuracy increased with the number of DD modules. This increasing property corresponds to the property in Taylor's expansion, as expected from the formula.

\begin{figure}[!t]
	\centering
	\includegraphics[width=\columnwidth]{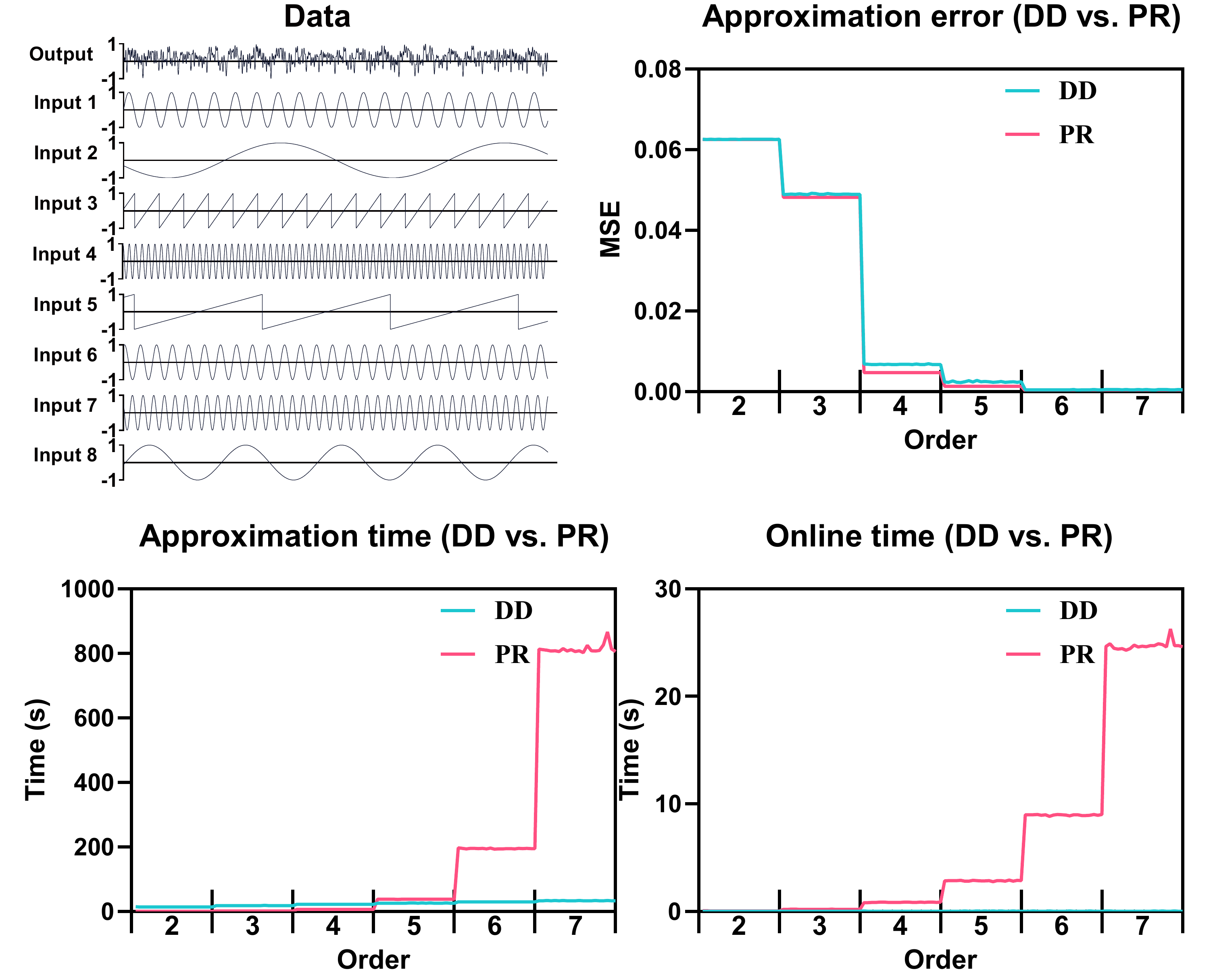}
	\caption{Comparison between DD and PR from 2 order to 7 order. Data: The size of inputs$\backslash$output:$ 1\times100000 $. DDs were trained for 1000000 epoch, and each epoch contains 2 data points. All algorithms for 20 runs in \textit{MATLAB 2019b}.	
	}
	\label{DDVSPR}
\end{figure}

This approximation property seems to be similar to polynomial regression. Thus this paper compares DD with polynomial regression. In order to directly evaluate the difference between DD and PR, we compared the approximate performance of 8 inputs to 1 output. The approximation data is shown in Fig. \ref{DDVSPR}. The simulation function is as follows.
\begin{equation} 
	\begin{aligned}
		O= &sin(I_{1})-cos(2I_{3})+sin(4I_{4}^{2})-I_{4}sin(5I_{5})+cos(8I_{6}^{2})\\&-I_{4}sin(I_{7})
		\label{eq.ddpr}
	\end{aligned}
\end{equation}
where $ O $ is Output, $ I_{i} $ denotes Input $ i $. The inputs and output were normalized to $ [-1,1] $ , respectively.  This paper compared the approximation error, the approximation time, and the online running time of the trained model ( forward-propagation time of calculating the output through the input data in \textit{MATLAB 2019b}).

Fig. \ref{DDVSPR} shows the comparative results. The approximation error of DD is larger than PR because of the suppression of higher-order terms by DD. However, as the order increases, all PR's approximation and online speed slow down rapidly, yet the DD does not change much. This means that DD is more suitable for online use in engineering, such as fitting sensor data and then running the model online.

\section{Regression}
Generally, a good algorithm should have considerable generalization capability. 

Using nine realworld datasets obtained from different fields, we compared DD with MLP architecture that imitated the cell body of neurons in \textit{MATLAB 2019b Neural Net toolbox} for regression.

\begin{figure*}[!t]
	\centering
	\includegraphics[width=1.7\columnwidth]{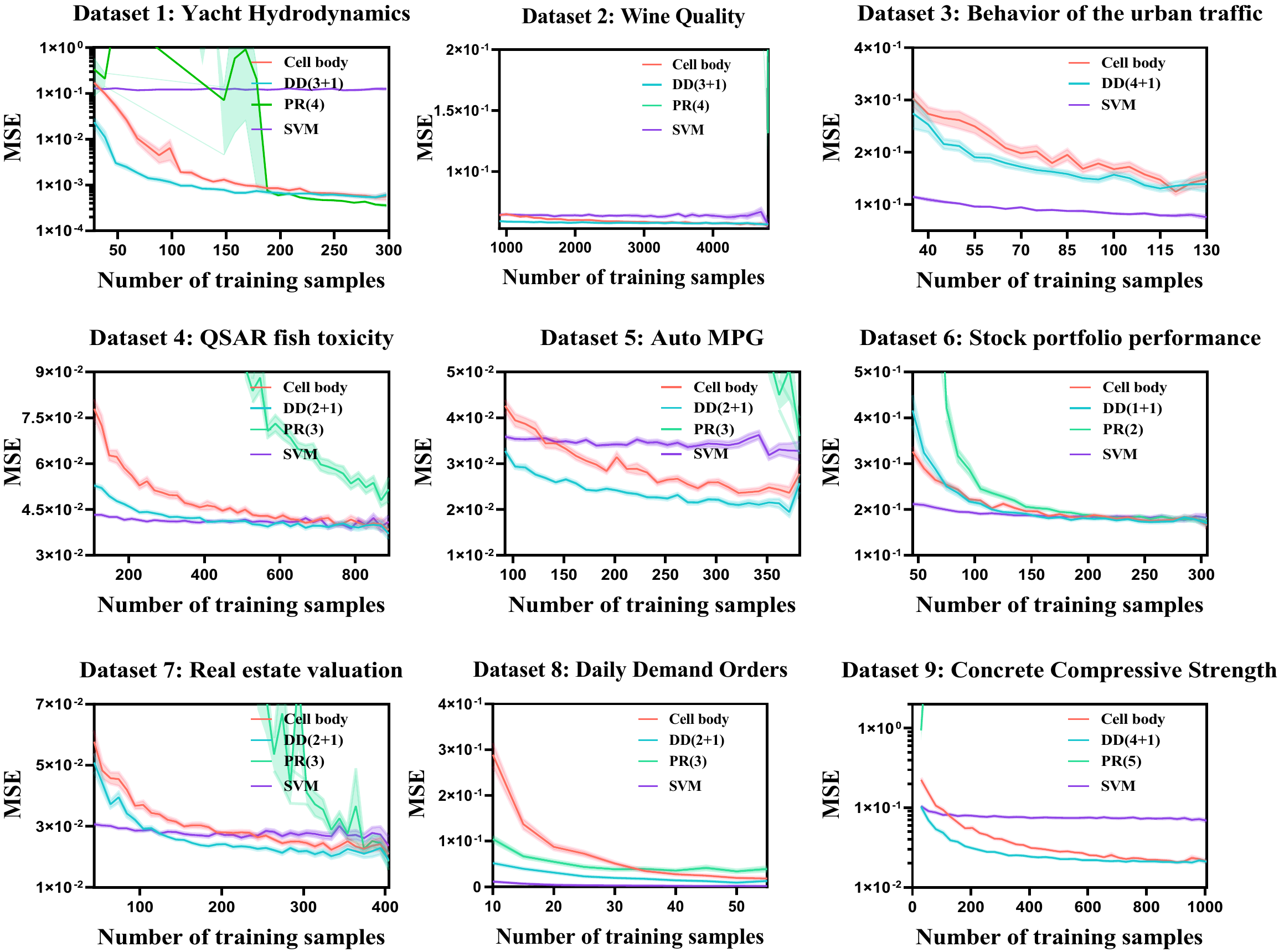}
	\caption{Comparison of generalization capability between Cell body Net, DD, PR, and SVM for regression problems. The order of PR is equal to the order of transformed DD. The inputs and outputs were normalized into [-1,1]. As an example, we selected $ tansig $ as the activation function of Cell body Net. Cell body Nets and DDs were run 100 times for all Datasets. SVMs were run 100 times for Dataset 1, 4-9 and were run 10 times for Dataset 2,3. PRs were run 100 times for Dataset 1, 4-9 and were run 10 times for Dataset 2 due to high time complexity of the algorithm. Furthermore, PR was failled for Dataset 3 due to high time complexity and space complexity(126113.4GB). MSE: $ mean\pm SEM $.}
	\label{fig10}
\end{figure*}

\begin{figure*}[!t]
	\centering
	\includegraphics[width=1.5\columnwidth]{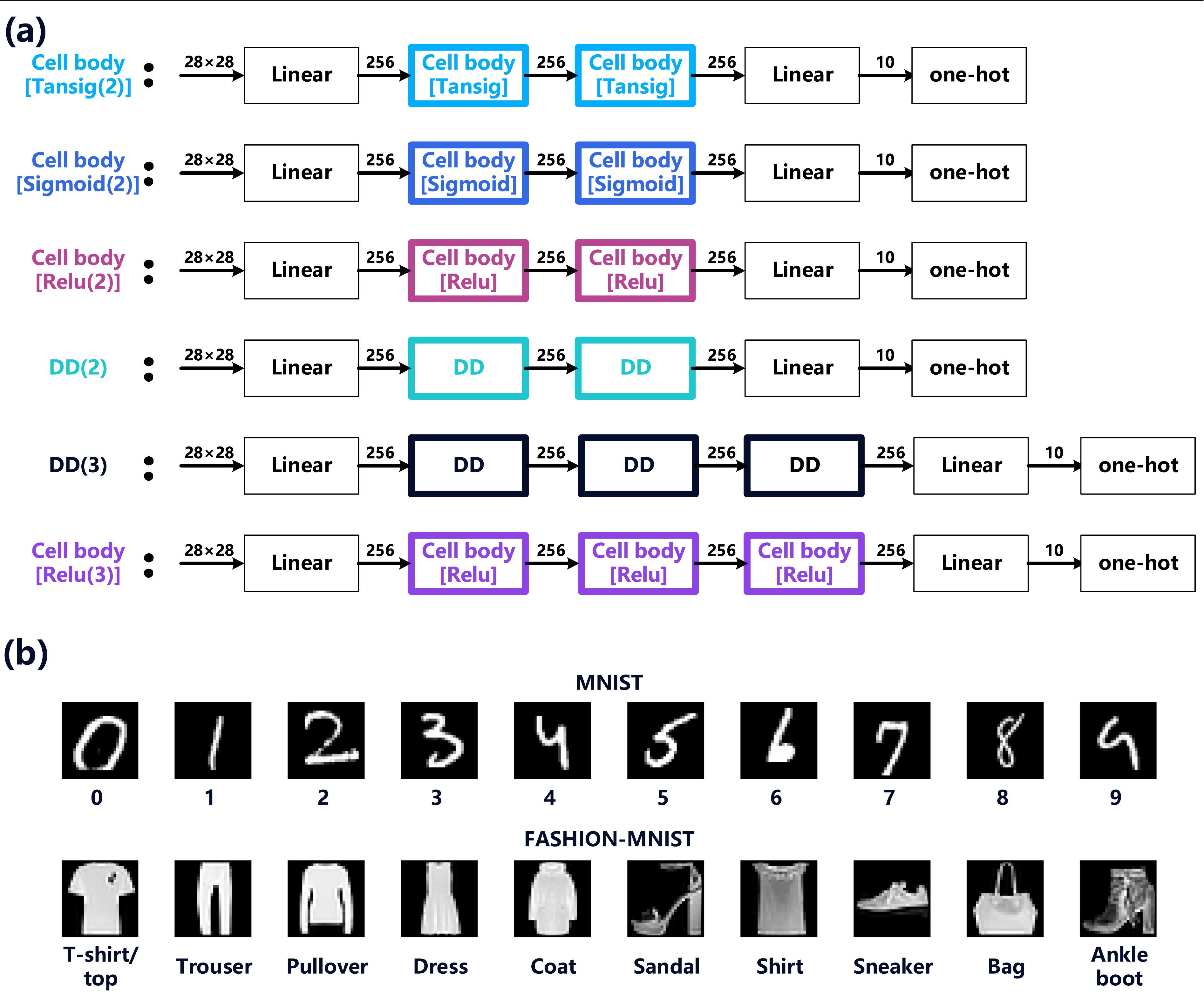}
	\caption{Classification experiments. (a) The cell body's and DD's models under similar architecture. (2) Examples from MNIST and FASHION-MNIST datasets.}
	\label{fig11}
\end{figure*}
\begin{figure*}[!t]
	\centering
	\includegraphics[width=1.5\columnwidth]{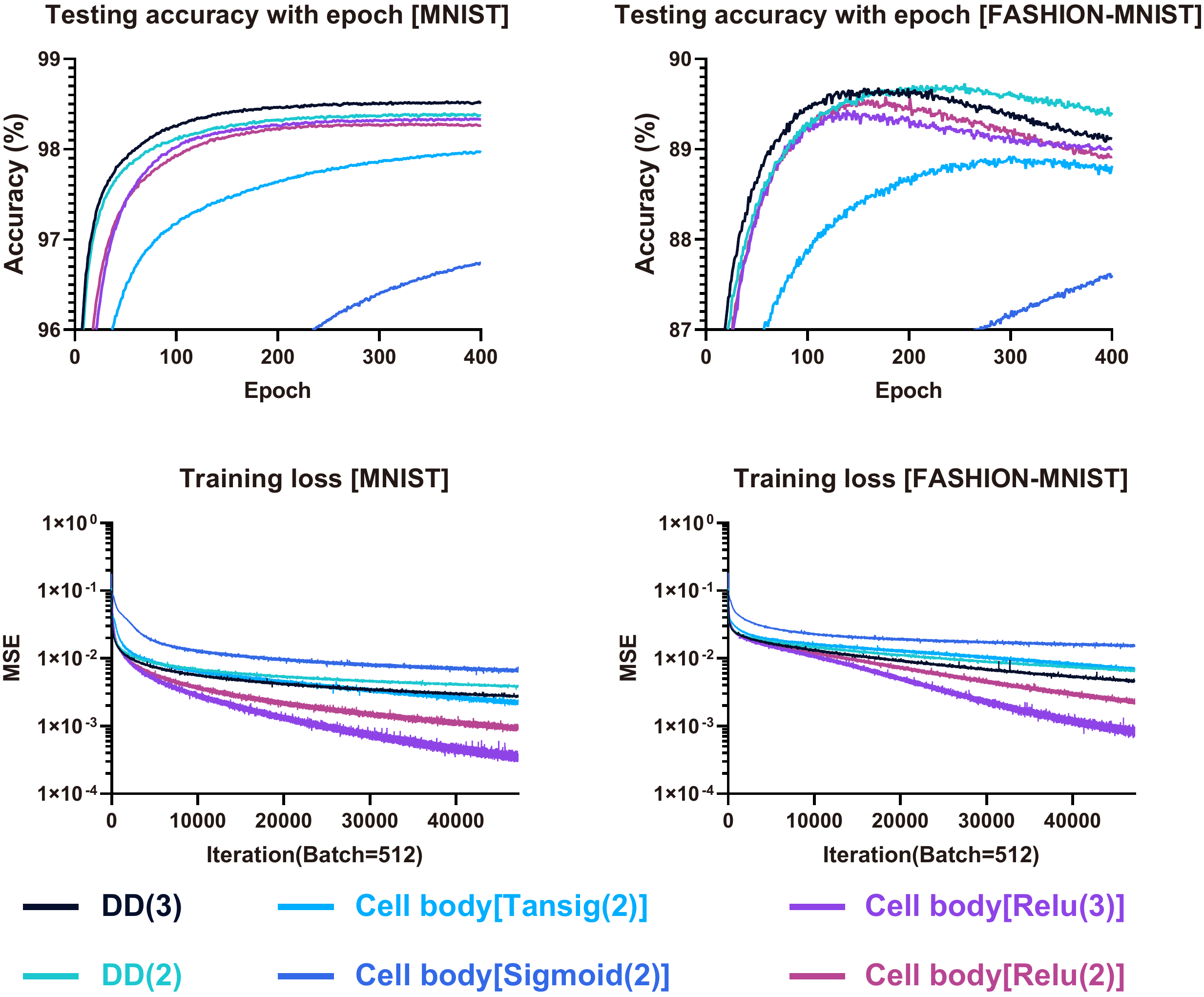}
	\caption{Comparison of generalization capability between Cell body Net and DD for classification problems. Batch size: 512. Optimizer:SGD. Learning rate: 0.05. Momentum: 0.9. All the networks for 20 runs.}
	\label{fig12}
\end{figure*}

\begin{figure*}[!t]
	\centering
	\includegraphics[width=1.7\columnwidth]{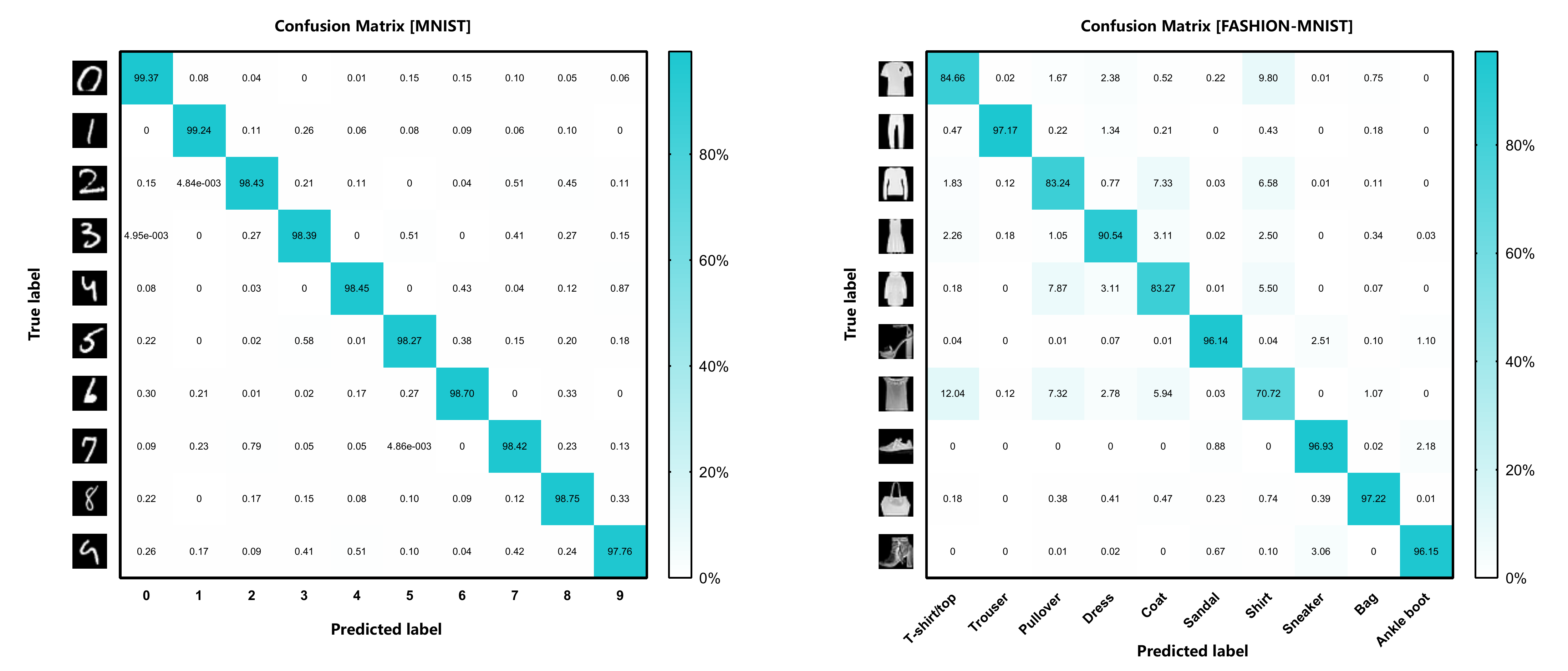}
	\caption{Confusion matrix for classification of MNIST and FASHION-MNIST using DD. All networks were run 20 times, and the network after each training classified 10,000 images from the test set. Therefore, 200,000 pictures were recognized for each dataset. MNIST: DD(3), Accuracy:98.59$\%$, Epoch:400, Batch size: 512. Optimizer:SGD. Learning rate: 0.05. Momentum: 0.9. FASHION-MNIST: DD(2), Accuracy:89.60$\% $, Epoch:255, Batch size: 512. }
	\label{confusion matrix}
\end{figure*}

\subsection{Datasets for regression}

\textit{See supplementary materials.}

\subsection{Experiments and results}

In practice, many essential factors potentially influence the test results of generalization capability for an algorithm and are critical to consider in our analyses. Here are some factors we try our best to consider.  Firstly, for Cell body Net, these factors include the number of neurons in the hidden layer, initial weights, activation function, and the training termination condition. Secondly, for DD, these factors are the number of DD modules and the training termination condition. Last but not least, in the aspect of data, these factors are the division of data, the number of training samples, and the distribution of the dataset itself.

First of all, we tested Cell body Net ten times to adjust training parameters for each dataset to be as comprehensive as possible. We found Cell body Net had better generalization capability for these datasets we obtained when the number of neurons in the hidden layer was 10. Besides, the training termination condition is to reach the preset training error (0.0001), maximum training times (5000), or validation checks (6).

Additionally, in order to increase the diversity of comparison, this paper also adds PR and SVM as comparison algorithms. Among them, SVM uses a linear kernel function, and the order of PR is equal to the order of transformed DD.

In this paper, we aim to explore the impact of as many data factors as possible on generalization capability, such as the number of training samples. Thus, we did not select cross-validation test methods used in previous literature (i.e., 5-fold cross-validation, 10-fold cross-validation or jackknife cross-validation test) and took the following exhaustive approach under the independent dataset test (\textit{See supplementary materials} \textbf{Algorithm: Testing the performance under different numbers of training samples. }) \cite{rodriguez2009sensitivity}.

Comparing Cell body Net with DD on various datasets, we found that DD gave results at least as good as, and often better than, Cell body Net (see Fig. \ref{fig10}).  Cell body Net looks like a black box. We have to adjust the hyperparameters over and over. DD is a ``white box''. The number of DD modules controls DD’s expression capacity. Excitingly, there is a one-to-one correspondence between the number of DD modules and precision from the outline to details. The white-box DD algorithm seems to get the model with excellent generalization capability more easily. In addition, DD and SVM have their own advantages and disadvantages for different datasets. For the same order, DD outperforms PR, especially in high order, which benefits from the learning rule of DD where the outline of output space is fitted by lower items first, and later the higher items modify the details (the suppression of high-order terms).

\section{Classification}

DD is the basic algorithm; thus, this paper mainly explores the fundamental properties of DD. We take the most commonly used MLP architecture as an example. The cell body's and dendritic modules' classification performances under similar architecture are explored on MNIST and FASHION-MNIST datasets (see Fig. \ref{fig11}).

\subsection{Datasets for classification}

This paper used Dataset 1: MNIST\cite{lecun1998gradient} and Dataset 2: FASHION-MNIST \cite{xiao2017fashion}. 
\textit{See supplementary materials.}

\subsection{Experiments and results}

Fig. \ref*{fig11}(a) shows the models in the experiment clearly. First, we compared the performance of the Cell body Net and DD when using the two modules. Among them, we analyzed Cell body Nets using different activation functions, respectively. Then, we explored the performance changes after adding a DD module. Meanwhile, we also added a module to the Cell body Net using the $ relu $ function for comparison.

Fig. \ref{fig12} shows the results of the experiment clearly. Exciting results only need a few words with firm evidence to explain. 

\textbf{(1) Interestingly, DDs show higher testing accuracy under greater training loss, which means better generalization capability.} In statistical learning theory, generalization originally refers to the model’s ability to generalize well the results obtained from the training set to a set of unseen samples drawn from the distribution same as that of the training set\cite{kushchu2002genetic,wang2014study}. The cell body has learned too many personal characteristics of the training set; thus, its loss to the training set is smaller, yet its loss to the test set is larger, which is over-fitting. Compared with the cell body, DD is easier to learn the common features in the data rather than the personalized features. Evidence: MNIST and FASHION-MNIST results.

\textbf{(2) DD shows the controllable training loss related to the number of DD modules, compared to the sustained decline of cell body’s training loss with iteration. } Evidence: Training loss of MNIST and FASHION-MNIST.

\textbf{(3) As the number of DD modules increases, training loss decreases.} Evidence: ``Training loss" in MNIST and FASHION-MNIST.

\textbf{(4) As the training loss decreases, testing accuracy increases for MNIST dataset.} Evidence: MNIST results.

\textbf{(5) We can improve DD’s generalization capability by adjusting the number of modules.} The complexity of the logical relationships within different data sets is different. DD's logical expression capacity can be effectively adjusted by the number of modules, which avoids over-fitting and generates a better generalization model. On the contrary, there is no one-to-one correspondence between Cell body Net’s logical expression capability and the number of modules. Thus, they are prone to over-fitting. Evidence: MNIST and FASHION-MNIST results.

\textbf{(6) DD shows faster effective convergence.} Evidence: ``Testing accuracy with epoch" in MNIST and FASHION-MNIST.

Additionally, Fig. \ref{confusion matrix} shows the confusion matrix for classification of MNIST and FASHION-MNIST using DD. The accuracy of MNIST is 98.59$ \% $, and the accuracy of FASHION-MNIST is 89.60$ \% $. The accuracy is higher than some typical basic machine learning algorithms (See TABLE \ref{Table1}).

\begin{table}[!t]
	\renewcommand{\arraystretch}{1.3}
	\caption{The comparison of test accuracy between DD with some basic algorithms }
	\label{table_example}
	\centering
	\begin{tabular}{|c||c||c|}	
		\hline
		\textbf{Classifier} & \textbf{MNIST} & \textbf{FASHION-MNIST}\\		
		\hline
		Linear Classifiers\cite{ministdata} & 92.40 $ \% $ &- \\
		\hline
		K-Nearest Neighbors (Euclidean)\cite{ministdata} & 97.60 $ \% $ &-\\
		\hline
		Boosted Stumps (17 leaves)\cite{ministdata} & 98.47 $ \% $ &-\\
		\hline
		40 PCA + quadratic classifier \cite{ministdata} & 96.70 $ \% $ &-\\
		\hline
		3-layer NN,500+150 HU \cite{ministdata} & 97.5 5$ \% $ &-\\
		\hline
		Convolutional net LeNet-1 \cite{ministdata} & 98.30$ \% $ &-\\
		\hline
		Decision Tree \cite{xiao2017fashion} & 88.60$ \% $ & 79.80 $ \% $\\
		\hline
		Extra Tree \cite{xiao2017fashion} & 84.70$ \% $ & 77.50 $ \% $\\
		\hline
		GaussianNB \cite{xiao2017fashion} & 52.40$ \% $ & 51.10 $ \% $\\
		\hline
		GradientBoosting \cite{xiao2017fashion} & 96.9$ \% $ & 88.0 $ \% $\\
		\hline
		k-Nearest Neighbors in \cite{xiao2017fashion} & 95.9$ \% $ & 85.4 $ \% $\\
		\hline
		LinearSVC \cite{xiao2017fashion} & 91.9$ \% $ & 83.6 $ \% $\\
		\hline
		Logistic Regression \cite{xiao2017fashion} & 91.7$ \% $ & 84.2 $ \% $ \\
		\hline
		MLP in \cite{xiao2017fashion} & 97.2$ \% $ & 87.1 $ \% $\\
		\hline
		Random Forest \cite{xiao2017fashion} & 97.0$ \% $ & 87.3 $ \% $ \\
		\hline
		SVC \cite{xiao2017fashion} & 97.3$ \% $ & 89.7 $ \% $ \\
		\hline
		\textbf{DD (our)}   & \textbf{98.59$ \% $} & \textbf{89.60 $ \% $} \\
		\hline
	\end{tabular}
	\label{Table1}
\end{table}

\section{Computational complexity}

\begin{figure}[!t]
	\centering
	\includegraphics[width=0.8\columnwidth]{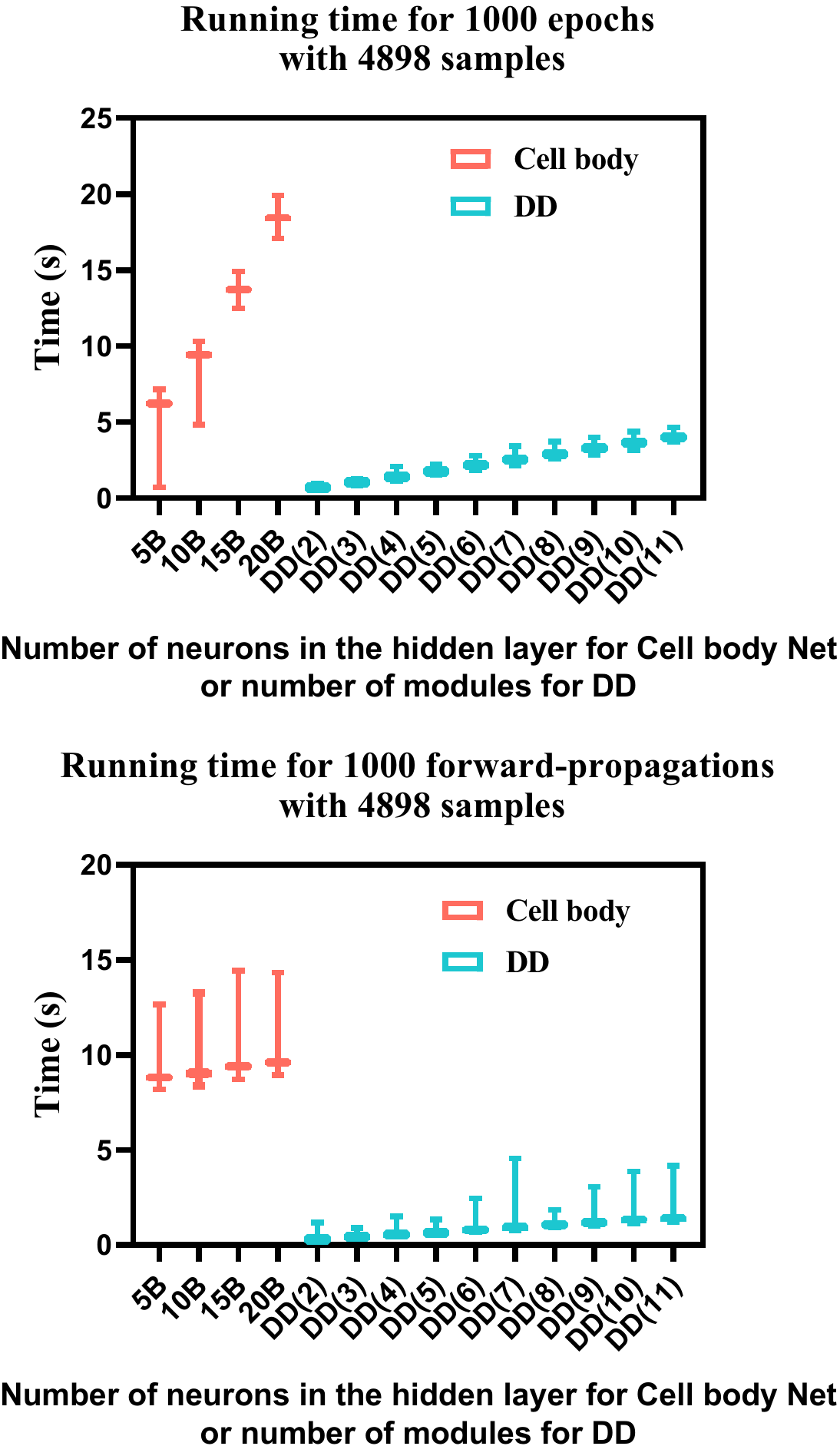}
	\caption{Comparison of running time in \textit{MATLAB 2019b} on a 2.2-GHz laptop PC. All the networks for 200 runs. ``$ xB $" expresses the three-layers Cell body Net with ``$ x $" neurons in the hidden layer. ``$ DD(y) $" expresses DD with ``$ y $" modules. For ``$ 5B $", the running time in forward-propagation is less than in the epoch. This is  because they are under different computer performance.}
	\label{fig13}
\end{figure}
\begin{figure}[!t]
	\centering
	\includegraphics[width=0.8\columnwidth]{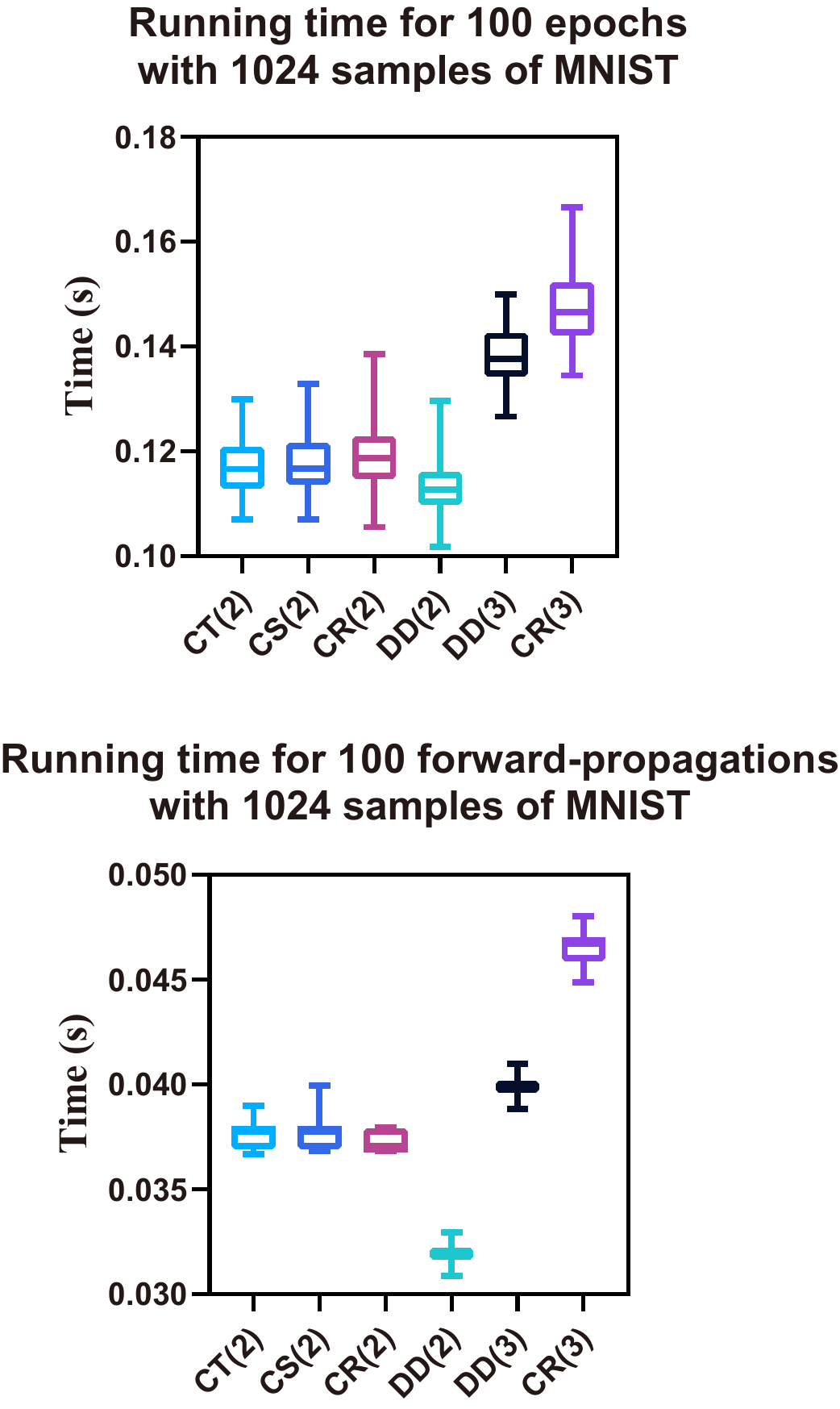}
	\caption{Comparison of running time in \textit{Python} using \textit{PyTorch (GPU)} and \textit{SGD optimizer} on a \textit{NVIDIA GeFore GTX 1060} and 2.2-GHz laptop PC. All the networks for 5900 runs. $ CT(2) $:Cell body [Tansig(2)]. $ CS(2) $:Cell body [Sigmoid(2)]. $ CR(2) $:Cell body [Relu(2)]. $ CR(3) $:Cell body[Relu(3)].}
	\label{fig14}
\end{figure}

The operation of DD only contains matrix multiplication and Hadamard product. It is well known that the computational complexity of Hadamard product is significantly lower than non-linear functions. Thus, the computational complexity of DD may be far lower than Cell body Net. To validate this conjecture, we designed two experiments about training networks and applications in real-time when using \textit{MATLAB} and \textit{PyTorch}.

Multiple factors influence the networks' running time or running speed (e.g., the learning rate and the termination condition for training networks). To eliminate these effects, we focus on the network's running time in the same number of epochs and forward propagations.

For \textit{MATLAB}, we tested the running time of Cell body Net and DD for 1000 epochs with 4898 samples from Wine Quality Dataset. One epoch is one forward-propagation plus one error-backpropagation for 4898 samples. In the aspect of an application in real-time, we tested the running time of Cell body Net and DD for 1000 forward-propagations with 4898 samples (see Fig. \ref{fig13}).

For \textit{PyTorch}, we tested the running time of Cell body Net and DD for 100 epochs with 1024 samples from MNIST dataset. One epoch is one forward-propagation plus one error-backpropagation for 1024 samples. In the aspect of an application in real-time, we tested the running time of Cell body Net and DD for 100 forward-propagation with 1024 samples (see Fig. \ref{fig14}).

Computer performance may vary between times. We wrote the models under the same testing conditions into a loop to ensure that they ran under similar computer performance as possible. Thus, the running speed of the models under the same conditions could be compared with each other. As we expected, DD's computational complexity was far lower than Cell body Net's, whether in an epoch or a forward-propagation.

\section{Additional Discussion}

\subsection{Definition of new concepts clearly}

\textbf{\textit{Dendrite Net or DD :}} Dendrite Net or DD refers to the whole fully connected network using DD modules and linear modules. As an analogy to ``MLP," MLP refers to the whole fully connected network using cell body modules. Because this paper is the first paper about DD modules, we did not embed DD modules to complex architecture(e.g., CNN, LSTM) and only explored the Dendrite Net.
Besides, considering the combination with cell body modules in the future, Dendrite Net also can be called DD (\textbf{D}en\textbf{d}rite).

\textbf{\textit{Cell body Net:} }Cell body Net refers to MLP.

\textbf{\textit{DD module:}} DD module refers to a ``$ WX\circ X $" module.

\textbf{\textit{Cell body module:}} Cell body module refers to a ``$ f(WX) $" module.

\textbf{\textit{Swish} and \textit{Mish} activation function\cite{2017arXiv171005941R,2019arXiv190808681M}:} \textit{Swish} is defined as $ f(x) =x \cdot sigmoid(x) $\cite{2017arXiv171005941R}, and \textit{Mish} is defined as $ f(x) =x \cdot tanh(softplus(x)) $\cite{2019arXiv190808681M}. They are activation function constructed by multiplying an input of the function and the output of traditional activation function. It is worth noting that the $ x $ in their definition refers to the input of the activation function, rather than the input of the entire network. Therefore, they are essentially an activation function. However, DD is with Hadamard product between the current inputs and the entire network's inputs and without activation function.

\textbf{\textit{DD's white-box attribute and ``interpretation" of other NNs:}}  ``Interpretation'' in other NNs is the ability to provide explanations in understandable terms to a human for a black-box model (e.g., deep neural networks) \cite{erhan2009visualizing, 2017arXiv170208608D,2018arXiv180200121Z,2018arXiv180610758H}.  However, the white-box attribute is different. All terms in the white-box model have the physical meaning that humans can understand. In other words, there are no parameters without physical meaning in the model, and the model is the interpretation itself. In order to assign the physical meaning to the parameters, the traditional white-box model is modeled based on physical properties \cite{lo2020identification,li2014review}. Nevertheless, modeling based on physical properties is one of the ways, but not the only one, to assign the physical meaning to the parameters.  Fourier transform decomposes the signal into components with periodic meaning. DD decomposes the complex system into independent components and interactive components (Contribution of input variables to results). The independent component and interactive components can be presented in a spectrum, just like the Fourier spectrum. Concretely, the white-box attribute refers to fully decompose an unknown composition into some simple components with physical meaning (e.g., Fourier transform and Fourier spectrum for decomposing signals) and read parameters of the white-box model, that is, simple components \cite{welch1967use}. In analogy to Fourier transform (trigonometric series means periodic components), the DD terms have physical meaning (the influence of inputs on the output). For example, DD terms contain $ x_{1} $ (first-order independent component), $ x_{1}^{2} $ (second-order independent component), and $ x_{1}x_{2} $ (second-order interactive component), etc ., where $ x_{1} $ and $ x_{2} $ are inputs. Similar to the result of Fourier transform, the application of DD's white-box attributes needs to be combined with specific engineering problems. For example, in this paper\cite{Liu2021}, we analyzed the brain's EEG-intent system using the white-box attributes. It is worth pointing out that, for classification tasks, it was shown as the contribution of the DD term for the corresponding class.

Summary of DD's attributes: DD inherits the gradual approximation properties of Taylor's expansion, DD has controllable precision. DD can generate a model with better generalization capability because of its controllable precision and learning rule where the outline of output space is fitted by lower items first, and later the higher items modify the details (the suppression of high-order terms). DD has lower computational complexity ($ O(2n-1) $ where $ n $ is the order of DD polynomial) because its operations only contain matrix multiplication and Hadamard product.

\subsection{Supplementary of theoretical proofs}

Weierstrass approximation theorem proved that DD could uniformly approximate any function that is merely continuous over a closed interval, and the unknown function need not be analytic (nor differentiable) \cite{farouki2012bernstein,stone1948generalized}. Thus, any function that is merely continuous over a closed interval can be approximated by the Dendrite Net after squashed to  $ [-1,1] $\cite{hornik1991approximation,hornik1989multilayer}.

\subsection{Classification and regression}

Classification and regression are the fundamental problems in many fields, such as fault diagnosis \cite{widodo2007support}, automation \cite{kehoe2015a,8491306,9216571}, computer vision (CV) \cite{szegedy2016rethinking}, and natural language processing (NLP) \cite{chapmanwendy2005classifying}.
Machine learning (ML) has been a useful tool to solve classification and regression problems \cite{ISI:000170489900001,quinlan1986induction,chang2011libsvm,fan2008liblinear,peduzzi1996simulation,geladi1986partial,belhumeur1997eigenfaces,hornik1989multilayer,wang2007naive,8809380,9091794,8432091,Liu2021a}. 
However, two key issues in ML algorithms remain to be resolved. 
(1) The existing ML algorithms only generate a black-box model \cite{ISI:000170489900001,quinlan1986induction,chang2011libsvm,peduzzi1996simulation,belhumeur1997eigenfaces,hornik1989multilayer,wang2007naive,9142403}.
(2) We cannot understand the detailed changes in the model's expression capacity and then achieve targeted precision, when tuning the hyperparameters. Thus, it is easy to cause over-fitting \cite{caruana2001overfitting}.

Classification is the task of dividing data according to sample features \cite{quinlan1986induction,chang2011libsvm,fan2008liblinear,peduzzi1996simulation,belhumeur1997eigenfaces,wang2007naive,guyon2002gene}. Therefore, it is natural to think that we can solve the problem by finding an appropriate classification curve or surface. However, ML algorithms using this strategy only generate a black-box model. Here, this paper presents a new strategy (DD). 
There is no doubt that the logical relationship (\textit{and$\backslash$or$\backslash$not}) among features determines the sample's class \cite{quinlan1990learning}. DD is to extract the amount of logical relationship information instead of finding a classification curve or surface. This paper designs the logical relationship expression (DD) with parameters to be solved and solve them. Then the sample can be classed with logical relationship expression according to many-valued logic theory \cite{rosser1957many,bolc2013many}.

Regression is the task of realizing the mapping between inputs (features) and outputs, and the outputs are continuous values rather than discrete classes \cite{chang2011libsvm,hornik1989multilayer,schielzeth2010simple}. In terms of interpretability, regression algorithms can be classified into black-box and white-box algorithms. White-box algorithms are more applicable to system identifications\cite{FITEEDD}. ML algorithms, such as SVM and MLP, are black-box \cite{chang2011libsvm,hornik1989multilayer}. Typical white-box algorithm is polynomial regression \cite{schielzeth2010simple,poggio1975optimal}. However, the computational complexity of PR does increase exponentially with the polynomial order. Here, \textbf{DD is the white-box algorithm with lower computational complexity.}

\subsection{Limitations}

As the first paper that proposed DD, this paper does not emphasize the application of DD to a specific field. We hope that DD will be applied to various basic engineering fields in the future. Because DD is a basic machine learning algorithm, this paper only explores the characteristics of general engineering. These characteristics are frequently used in the engineering field. In fact, in general engineering, the typical BP neural network (i.e., MLP), SVM, and other small and simple algorithms rather than large-scale algorithms are commonly used at present. For some special field data sets, such as image recognition data sets (e.g., CIFAR-10, CIFAR-100, COCO, and Places365-Challenge), this paper did not study them. CIFAR-10, CIFAR-100, COCO, and Places365-Challenge are the datasets of color images; thus, special processing in the network is required. Simply using basic networks (e.g., DD, SVM, MLP) is generally inferior to private networks (e.g., AlexNet \cite{krizhevsky2012imagenet}). For general engineering, the input data dimension (784) and the number of categories of MNIST and FASHION-MNIST are generally relatively larger than engineering datasets, such as the data set composed of various sensors in robot control. In the future, we will improve the DD module and use it in special fields such as image recognition \cite{Liu2020}. Based on maintaining its advantages, extending DD to a deeper model with more complex structures will be further studied in the future \cite{Liu2020}.

\begin{figure}[!t]
	\centering
	\includegraphics[width=0.8\columnwidth]{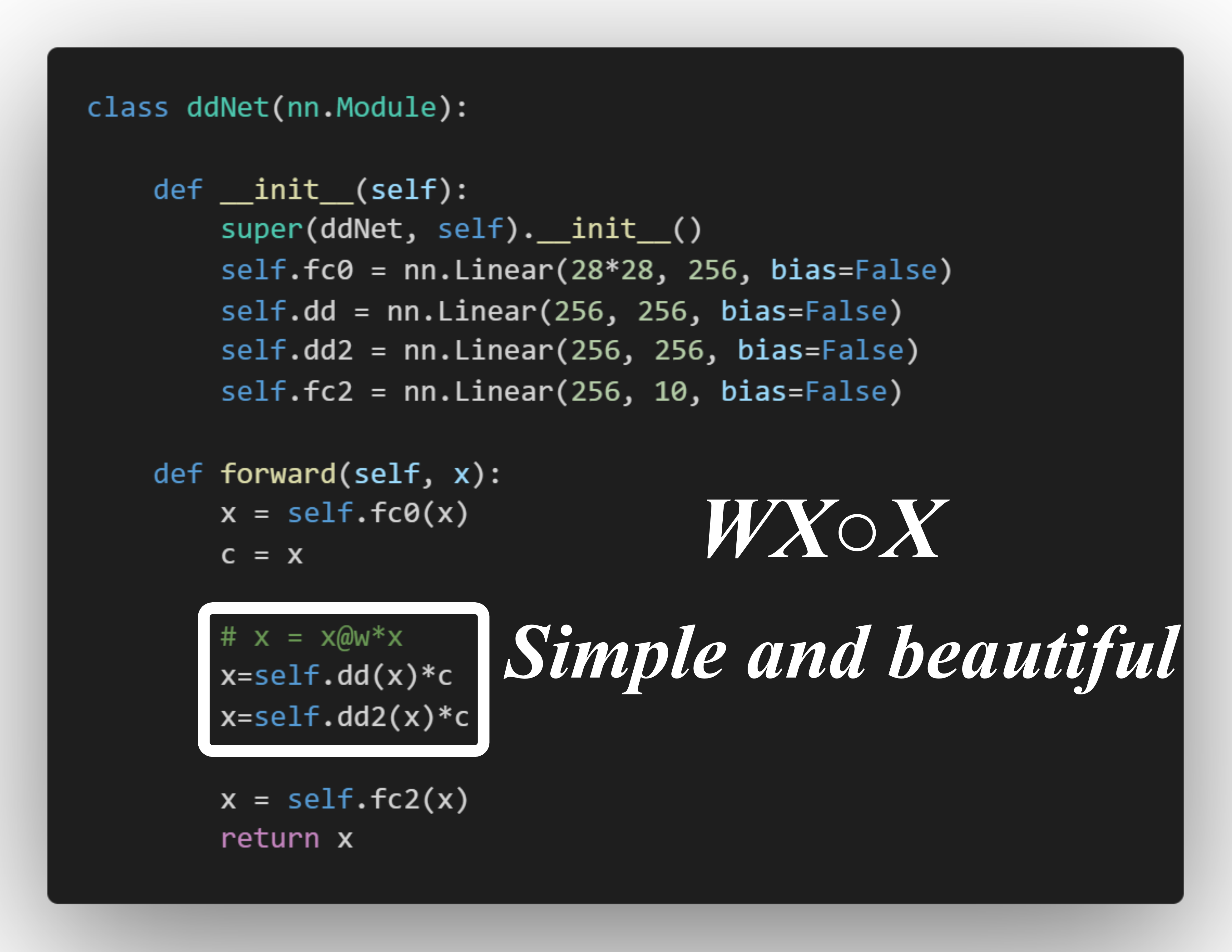}
	\caption{Example of DD, Dendrite Net or DD Net (\textit{Python} code). DD can be seen as a module plugged into other large networks, or as an independent neural network for use alone. Its use can be understood by analogy with MLP or Cell body which can be used alone or inserted into other large networks.	
	}
	\label{fig15}
\end{figure}

\section{Conclusion}

In this paper, a basic machine learning algorithm, named Dendrite Net or DD, is proposed. DD is a white-box ML algorithm for classification, regression, and system identification. DD aims to design the logical expression among inputs with controllable precision.

We highlight DD’s white-box attribute, controllable precision for better generalization capability, and lower computational complexity. Needless to say, the experiment results are exciting, DD is open source, and everyone can verify these. Additionally, for basic algorithms, many factors can affect performance. This paper proved the nature of DD by comparison under the same conditions from the basic definition.

DD module is simple and beautiful. In the future, DD is not only used for generalized engineering as other basic ML algorithms but also has vast development potential as a module for deep learning. A combination of Dendrite Net and Cell body Net maybe improve the present artificial neuron or ANNs \cite{Liu2020}.

\section*{Acknowledgments}
Gang Liu presented DD. Jing Wang offered advice. The authors would like to thank the researchers who contacted us, discussed with us, and offered suggestions for our preprint. The authors sincerely thank the editor and six anonymous reviewers for their valuable comments on improving the paper. These suggestions improved this manuscript.


\bibliographystyle{IEEEtran}

\bibliography{ArXiv-DD}

\begin{thebibliography}{10}
\providecommand{\url}[1]{#1}
\csname url@samestyle\endcsname
\providecommand{\newblock}{\relax}
\providecommand{\bibinfo}[2]{#2}
\providecommand{\BIBentrySTDinterwordspacing}{\spaceskip=0pt\relax}
\providecommand{\BIBentryALTinterwordstretchfactor}{4}
\providecommand{\BIBentryALTinterwordspacing}{\spaceskip=\fontdimen2\font plus
\BIBentryALTinterwordstretchfactor\fontdimen3\font minus
  \fontdimen4\font\relax}
\providecommand{\BIBforeignlanguage}[2]{{%
\expandafter\ifx\csname l@#1\endcsname\relax
\typeout{** WARNING: IEEEtran.bst: No hyphenation pattern has been}%
\typeout{** loaded for the language `#1'. Using the pattern for}%
\typeout{** the default language instead.}%
\else
\language=\csname l@#1\endcsname
\fi
#2}}
\providecommand{\BIBdecl}{\relax}
\BIBdecl

\bibitem{mcculloch1943logical}
W.~S. McCulloch and W.~Pitts, ``A logical calculus of the ideas immanent in
  nervous activity,'' \emph{The bulletin of mathematical biophysics}, vol.~5,
  no.~4, pp. 115--133, 1943.

\bibitem{gidon2020dendritic}
A.~Gidon, T.~A. Zolnik, P.~Fidzinski, F.~Bolduan, A.~Papoutsi, P.~Poirazi,
  M.~Holtkamp, I.~Vida, and M.~E. Larkum, ``Dendritic action potentials and
  computation in human layer 2/3 cortical neurons,'' \emph{Science}, vol. 367,
  no. 6473, pp. 83--87, 2020.

\bibitem{london2005dendritic}
M.~London and M.~H{\"a}usser, ``Dendritic computation,'' \emph{Annu. Rev.
  Neurosci.}, vol.~28, pp. 503--532, 2005.

\bibitem{mel1994information}
B.~W. Mel, ``Information processing in dendritic trees,'' \emph{Neural
  computation}, vol.~6, no.~6, pp. 1031--1085, 1994.

\bibitem{shepherd1987logic}
G.~M. Shepherd and R.~K. Brayton, ``Logic operations are properties of
  computer-simulated interactions between excitable dendritic spines,''
  \emph{Neuroscience}, vol.~21, no.~1, pp. 151--165, 1987.

\bibitem{tang1998learning}
Z.~Tang, Q.-p. Cao, and O.~Ishizuka, ``A learning multiple-valued logic
  network: Algebra, algorithm, and applications,'' \emph{IEEE transactions on
  computers}, vol.~47, no.~2, pp. 247--251, 1998.

\bibitem{todo2019neurons}
Y.~Todo, Z.~Tang, H.~Todo, J.~Ji, and K.~Yamashita, ``Neurons with
  multiplicative interactions of nonlinear synapses,'' \emph{International
  journal of neural systems}, vol.~29, no.~08, p. 1950012, 2019.

\bibitem{sun2018bi}
J.~Sun, S.~Gao, H.~Dai, J.~Cheng, M.~Zhou, and J.~Wang, ``Bi-objective elite
  differential evolution algorithm for multivalued logic networks,'' \emph{IEEE
  Transactions on Cybernetics}, vol.~50, no.~1, pp. 233--246, 2018.

\bibitem{gao2018dendritic}
S.~Gao, M.~Zhou, Y.~Wang, J.~Cheng, H.~Yachi, and J.~Wang, ``Dendritic neuron
  model with effective learning algorithms for classification, approximation,
  and prediction,'' \emph{IEEE transactions on neural networks and learning
  systems}, vol.~30, no.~2, pp. 601--614, 2019.

\bibitem{durbin1989product}
R.~Durbin and D.~E. Rumelhart, ``Product units: A computationally powerful and
  biologically plausible extension to backpropagation networks,'' \emph{Neural
  computation}, vol.~1, no.~1, pp. 133--142, 1989.

\bibitem{bolc2013many}
L.~Bolc and P.~Borowik, \emph{Many-valued logics 1: theoretical
  foundations}.\hskip 1em plus 0.5em minus 0.4em\relax Springer Science \&
  Business Media, 2013.

\bibitem{rosser1957many}
J.~B. Rosser, \emph{Many-valued logics}.\hskip 1em plus 0.5em minus 0.4em\relax
  United States Air Force, Office of Scientific Research, 1957.

\bibitem{hao2009arithmetic}
J.~Hao, X.-d. Wang, Y.~Dan, M.-m. Poo, and X.-h. Zhang, ``An arithmetic rule
  for spatial summation of excitatory and inhibitory inputs in pyramidal
  neurons,'' \emph{Proceedings of the National Academy of Sciences}, vol. 106,
  no.~51, pp. 21\,906--21\,911, 2009.

\bibitem{quinlan1990learning}
J.~R. Quinlan, ``Learning logical definitions from relations,'' \emph{Machine
  learning}, vol.~5, no.~3, pp. 239--266, 1990.

\bibitem{Liu2020}
\BIBentryALTinterwordspacing
G.~Liu, ``{It may be time to improve the neuron of artificial neural
  network},'' 6 2020. [Online]. Available:
  \url{https://doi.org/10.36227/techrxiv.12477266}
\BIBentrySTDinterwordspacing

\bibitem{vaswani2017attention}
A.~Vaswani, N.~Shazeer, N.~Parmar, J.~Uszkoreit, L.~Jones, A.~N. Gomez,
  {\L}.~Kaiser, and I.~Polosukhin, ``Attention is all you need,'' in
  \emph{Advances in neural information processing systems}, 2017, pp.
  5998--6008.

\bibitem{ma2005constructive}
L.~Ma and K.~Khorasani, ``Constructive feedforward neural networks using
  hermite polynomial activation functions,'' \emph{IEEE Transactions on Neural
  Networks}, vol.~16, no.~4, pp. 821--833, 2005.

\bibitem{jiang2016potential}
B.~Jiang, J.~Li, and H.~Guo, ``Potential energy surfaces from high fidelity
  fitting of ab initio points: the permutation invariant polynomial-neural
  network approach,'' \emph{International Reviews in Physical Chemistry},
  vol.~35, no.~3, pp. 479--506, 2016.

\bibitem{schielzeth2010simple}
H.~Schielzeth, ``Simple means to improve the interpretability of regression
  coefficients,'' \emph{Methods in Ecology and Evolution}, vol.~1, no.~2, pp.
  103--113, 2010.

\bibitem{poggio1975optimal}
T.~Poggio, ``On optimal nonlinear associative recall,'' \emph{Biological
  Cybernetics}, vol.~19, no.~4, pp. 201--209, 1975.

\bibitem{rumelhart1986learning}
D.~E. Rumelhart, G.~E. Hinton, and R.~J. Williams, ``Learning representations
  by back-propagating errors,'' \emph{nature}, vol. 323, no. 6088, pp.
  533--536, 1986.

\bibitem{rodriguez2009sensitivity}
J.~D. Rodriguez, A.~Perez, and J.~A. Lozano, ``Sensitivity analysis of k-fold
  cross validation in prediction error estimation,'' \emph{IEEE transactions on
  pattern analysis and machine intelligence}, vol.~32, no.~3, pp. 569--575,
  2009.

\bibitem{lecun1998gradient}
Y.~LeCun, L.~Bottou, Y.~Bengio, and P.~Haffner, ``Gradient-based learning
  applied to document recognition,'' \emph{Proceedings of the IEEE}, vol.~86,
  no.~11, pp. 2278--2324, 1998.

\bibitem{xiao2017fashion}
H.~Xiao, K.~Rasul, and R.~Vollgraf, ``Fashion-mnist: a novel image dataset for
  benchmarking machine learning algorithms,'' \emph{arXiv preprint
  arXiv:1708.07747}, 2017.

\bibitem{kushchu2002genetic}
I.~Kushchu, ``Genetic programming and evolutionary generalization,'' \emph{IEEE
  transactions on Evolutionary Computation}, vol.~6, no.~5, pp. 431--442, 2002.

\bibitem{wang2014study}
X.-Z. Wang, H.-J. Xing, Y.~Li, Q.~Hua, C.-R. Dong, and W.~Pedrycz, ``A study on
  relationship between generalization abilities and fuzziness of base
  classifiers in ensemble learning,'' \emph{IEEE Transactions on Fuzzy
  Systems}, vol.~23, no.~5, pp. 1638--1654, 2014.

\bibitem{ministdata}
C.~C. Yann~LeCun and C.~J. Burges, ``The mnist dataset of handwritten digits,''
  \url{http://yann.lecun.com/exdb/mnist/}, accessed April 4, 2010.

\bibitem{2017arXiv171005941R}
P.~{Ramachandran}, B.~{Zoph}, and Q.~V. {Le}, ``{Searching for Activation
  Functions},'' \emph{arXiv e-prints}, p. arXiv:1710.05941, Oct. 2017.

\bibitem{2019arXiv190808681M}
D.~{Misra}, ``{Mish: A Self Regularized Non-Monotonic Activation Function},''
  \emph{arXiv e-prints}, p. arXiv:1908.08681, Aug. 2019.

\bibitem{erhan2009visualizing}
D.~Erhan, Y.~Bengio, A.~Courville, and P.~Vincent, ``Visualizing higher-layer
  features of a deep network,'' \emph{University of Montreal}, vol. 1341,
  no.~3, p.~1, 2009.

\bibitem{2017arXiv170208608D}
F.~{Doshi-Velez} and B.~{Kim}, ``{Towards A Rigorous Science of Interpretable
  Machine Learning},'' \emph{arXiv e-prints}, p. arXiv:1702.08608, Feb. 2017.

\bibitem{2018arXiv180200121Z}
Q.~{Zhang}, Y.~{Yang}, H.~{Ma}, and Y.~{Nian Wu}, ``{Interpreting CNNs via
  Decision Trees},'' \emph{arXiv e-prints}, p. arXiv:1802.00121, Jan. 2018.

\bibitem{2018arXiv180610758H}
S.~{Hooker}, D.~{Erhan}, P.-J. {Kindermans}, and B.~{Kim}, ``{A Benchmark for
  Interpretability Methods in Deep Neural Networks},'' \emph{arXiv e-prints},
  p. arXiv:1806.10758, Jun. 2018.

\bibitem{lo2020identification}
O.~Lo-Thong, P.~Charton, X.~F. Cadet, B.~Grondin-Perez, E.~Saavedra, C.~Damour,
  and F.~Cadet, ``Identification of flux checkpoints in a metabolic pathway
  through white-box, grey-box and black-box modeling approaches,''
  \emph{Scientific reports}, vol.~10, no.~1, pp. 1--19, 2020.

\bibitem{li2014review}
X.~Li and J.~Wen, ``Review of building energy modeling for control and
  operation,'' \emph{Renewable and Sustainable Energy Reviews}, vol.~37, pp.
  517--537, 2014.

\bibitem{welch1967use}
P.~Welch, ``The use of fast fourier transform for the estimation of power
  spectra: a method based on time averaging over short, modified
  periodograms,'' \emph{IEEE Transactions on audio and electroacoustics},
  vol.~15, no.~2, pp. 70--73, 1967.

\bibitem{Liu2021}
\BIBentryALTinterwordspacing
G.~Liu and J.~Wang, ``{EEGG: An analytic brain-computer interface algorithm},''
  1 2021. [Online]. Available: \url{https://doi.org/10.36227/techrxiv.13516145}
\BIBentrySTDinterwordspacing

\bibitem{farouki2012bernstein}
R.~T. Farouki, ``The bernstein polynomial basis: A centennial retrospective,''
  \emph{Computer Aided Geometric Design}, vol.~29, no.~6, pp. 379--419, 2012.

\bibitem{stone1948generalized}
M.~H. Stone, ``The generalized weierstrass approximation theorem,''
  \emph{Mathematics Magazine}, vol.~21, no.~5, pp. 237--254, 1948.

\bibitem{hornik1991approximation}
K.~Hornik, ``Approximation capabilities of multilayer feedforward networks,''
  \emph{Neural networks}, vol.~4, no.~2, pp. 251--257, 1991.

\bibitem{hornik1989multilayer}
K.~Hornik, M.~Stinchcombe, H.~White \emph{et~al.}, ``Multilayer feedforward
  networks are universal approximators.'' \emph{Neural networks}, vol.~2,
  no.~5, pp. 359--366, 1989.

\bibitem{widodo2007support}
A.~Widodo and B.~Yang, ``Support vector machine in machine condition monitoring
  and fault diagnosis,'' \emph{Mechanical Systems and Signal Processing},
  vol.~21, no.~6, pp. 2560--2574, 2007.

\bibitem{kehoe2015a}
B.~Kehoe, S.~Patil, P.~Abbeel, and K.~Goldberg, ``A survey of research on cloud
  robotics and automation,'' \emph{IEEE Transactions on Automation Science and
  Engineering}, vol.~12, no.~2, pp. 398--409, 2015.

\bibitem{8491306}
G.~{Wang}, J.~{Lu}, K.~{Choi}, and G.~{Zhang}, ``A transfer-based additive
  ls-svm classifier for handling missing data,'' \emph{IEEE Transactions on
  Cybernetics}, vol.~50, no.~2, pp. 739--752, 2020.

\bibitem{9216571}
Z.~{Ma} and P.~{Huang}, ``Adaptive neural-network controller for an uncertain
  rigid manipulator with input saturation and full-order state constraint,''
  \emph{IEEE Transactions on Cybernetics}, pp. 1--9, 2020.

\bibitem{szegedy2016rethinking}
C.~Szegedy, V.~Vanhoucke, S.~Ioffe, J.~Shlens, and Z.~Wojna, ``Rethinking the
  inception architecture for computer vision,'' pp. 2818--2826, 2016.

\bibitem{chapmanwendy2005classifying}
W.~Chapmanwendy, M.~Christensenlee, M.~Wagnermichael, J.~Haugpeter, Ivanovoleg,
  N.~Dowlingjohn, and T.~Olszewskirobert, ``Classifying free-text triage chief
  complaints into syndromic categories with natural languages processing,''
  \emph{Artificial Intelligence in Medicine}, 2005.

\bibitem{ISI:000170489900001}
L.~Breiman, ``\BIBforeignlanguage{{English}}{{Random forests}},''
  \emph{\BIBforeignlanguage{{English}}{{MACHINE LEARNING}}}, vol.~{45},
  no.~{1}, pp. {5--32}, {OCT} {2001}.

\bibitem{quinlan1986induction}
J.~R. Quinlan, ``Induction of decision trees,'' \emph{Machine learning},
  vol.~1, no.~1, pp. 81--106, 1986.

\bibitem{chang2011libsvm}
C.-C. Chang and C.-J. Lin, ``Libsvm: A library for support vector machines,''
  \emph{ACM transactions on intelligent systems and technology (TIST)}, vol.~2,
  no.~3, pp. 1--27, 2011.

\bibitem{fan2008liblinear}
R.-E. Fan, K.-W. Chang, C.-J. Hsieh, X.-R. Wang, and C.-J. Lin, ``Liblinear: A
  library for large linear classification,'' \emph{Journal of machine learning
  research}, vol.~9, no. Aug, pp. 1871--1874, 2008.

\bibitem{peduzzi1996simulation}
P.~Peduzzi, J.~Concato, E.~Kemper, T.~R. Holford, and A.~R. Feinstein, ``A
  simulation study of the number of events per variable in logistic regression
  analysis,'' \emph{Journal of clinical epidemiology}, vol.~49, no.~12, pp.
  1373--1379, 1996.

\bibitem{geladi1986partial}
P.~Geladi and B.~R. Kowalski, ``Partial least-squares regression: a tutorial,''
  \emph{Analytica chimica acta}, vol. 185, pp. 1--17, 1986.

\bibitem{belhumeur1997eigenfaces}
P.~N. Belhumeur, J.~P. Hespanha, and D.~J. Kriegman, ``Eigenfaces vs.
  fisherfaces: Recognition using class specific linear projection,'' \emph{IEEE
  Transactions on pattern analysis and machine intelligence}, vol.~19, no.~7,
  pp. 711--720, 1997.

\bibitem{wang2007naive}
Q.~Wang, G.~M. Garrity, J.~M. Tiedje, and J.~R. Cole, ``Naive bayesian
  classifier for rapid assignment of rrna sequences into the new bacterial
  taxonomy,'' \emph{Applied and environmental microbiology}, vol.~73, no.~16,
  pp. 5261--5267, 2007.

\bibitem{8809380}
J.~{Ma}, H.~{Zhang}, and T.~W.~S. {Chow}, ``Multilabel classification with
  label-specific features and classifiers: A coarse- and fine-tuned
  framework,'' \emph{IEEE Transactions on Cybernetics}, pp. 1--15, 2019.

\bibitem{9091794}
S.~{Liu}, L.~{Wang}, B.~{Yang}, J.~{Zhou}, Z.~{Chen}, and H.~{Dong},
  ``Improvement of neural-network classifiers using fuzzy floating centroids,''
  \emph{IEEE Transactions on Cybernetics}, pp. 1--13, 2020.

\bibitem{8432091}
S.~{Feng} and C.~L.~P. {Chen}, ``Fuzzy broad learning system: A novel
  neuro-fuzzy model for regression and classification,'' \emph{IEEE
  Transactions on Cybernetics}, vol.~50, no.~2, pp. 414--424, 2020.

\bibitem{Liu2021a}
G.~Liu, L.~Wang, and J.~Wang, ``A novel energy-motion model for continuous
  {sEMG} decoding: from muscle energy to motor pattern,'' vol.~18, no.~1, p.
  016019, feb 2021.

\bibitem{9142403}
Q.~{Chen}, B.~{Xue}, and M.~{Zhang}, ``Rademacher complexity for enhancing the
  generalization of genetic programming for symbolic regression,'' \emph{IEEE
  Transactions on Cybernetics}, pp. 1--14, 2020.

\bibitem{caruana2001overfitting}
R.~Caruana, S.~Lawrence, and C.~L. Giles, ``Overfitting in neural nets:
  Backpropagation, conjugate gradient, and early stopping,'' in \emph{Advances
  in neural information processing systems}, 2001, pp. 402--408.

\bibitem{guyon2002gene}
I.~Guyon, J.~Weston, S.~Barnhill, and V.~Vapnik, ``Gene selection for cancer
  classification using support vector machines,'' \emph{Machine learning},
  vol.~46, no. 1-3, pp. 389--422, 2002.

\bibitem{FITEEDD}
J.~W. Gang~Liu, ``A relation spectrum inheriting taylor series: muscle synergy
  and coupling for hand,'' \emph{Frontiers of Information Technology \&
  Electronic Engineering}, in press,2021.

\bibitem{krizhevsky2012imagenet}
A.~Krizhevsky, I.~Sutskever, and G.~E. Hinton, ``Imagenet classification with
  deep convolutional neural networks,'' \emph{Advances in neural information
  processing systems}, vol.~25, pp. 1097--1105, 2012.

\end{thebibliography}

%
%
%

\newpage

\vfill

\cleardoublepage



\section*{Supplementary material}

# Dendrite Net: A White-Box Module for Classification, Regression, and System Identification (Supplementary materials)

Gang Liu, Jing Wang

*Abstract*—This document is the supplementary materials for *Dendrite Net: A White-Box Module for Classification, Regression, and System Identification*. DD code is available at *GitHub:Gang neuron*.

*Index Terms*—Machine learning, algorithms, engineering, artificial intelligence, pattern recognition.

## I. RELATED WORK

### A. Basic ML Algorithms

So far, some basic ML algorithms have been proposed. **However, there is no DD.** This section briefly reviews these basic algorithms.

**Least Squares Regression [1]:** The least-squares method is a statistical procedure to find the best fit for a set of data points by minimizing the sum of the offsets or residuals of points from the plotted curve. It is a typical method for performing linear regression.

**Logistic Regression [2]:** Logistic regression measures the relationship between the categorical dependent variable and one or more independent variables by estimating probabilities using a logistic function, which is the cumulative distribution function of logistic distribution.

**Linear discriminant analysis (LDA) [3]:** LDA is a discriminant approach that attempts to model differences among samples assigned to certain groups. The method aims to maximize the ratio of the between-group variance and the within-group variance.

**Decision Trees [4]:** A decision tree is a decision support tool that uses a tree-like graph or model of decisions and possible consequences, including chance-event outcomes, resource costs, and utility.

**Naive Bayes Classification [5]:** Naive Bayes classifiers are a family of simple probabilistic classifiers based on applying Bayes' theorem with strong (naive) independence assumptions between the features.

**Support Vector Machines [6]:** A support vector machine constructs a hyperplane or set of hyperplanes in a high- or infinite-dimensional space, which can be used for classification, regression, or other tasks like outliers detection.

**Ensemble Methods [7]:** Ensemble methods are learning algorithms that construct a set of classifiers and then classify new data points by taking a weighted vote of their predictions. The original ensemble method is Bayesian averaging, but more recent algorithms include error-correcting output coding, bagging, and boosting.

**Clustering Algorithms [8]:** Clustering is the task of grouping a set of objects such that objects in the same group (cluster) are more similar to each other than to those in other groups (e.g., k-means clustering algorithm).

**Random forests or random decision forests [9]:** These are an ensemble learning method for classification, regression, and other tasks that operate by constructing a multitude of decision trees at training time and outputting the class that is the mode of the classes (classification) or mean prediction (regression) of the individual trees.

**Artificial Neural Networks (ANN) [10]:** An ANN is based on a collection of connected units or nodes called artificial neurons, which loosely model the neurons in a biological brain. However, the existing ANN does not consider dendrite's logical operation. From a particular perspective, the existing neural networks also have the fuzzy concept of information extraction. Nevertheless, this is a fuzzy non-linear mapping, not a clear, logical relationship extractor.

In addition to the above typical basic algorithms, some algorithms, such as memristive neural networks [11], are combined with the application environment. These will not be covered here. It is worth emphasizing that the above algorithms have their own advantages in different aspects; however, there is no DD. The benefits of DD, such as white-box attribute, controllable precision for better generalization capability, and lower computational complexity, might pose new changes in many fields in the future.

## II. DD

### A. Theoretical proofs

Two points need to be proved: (1) the ability of DD to approximate any continuous function, to any desired accuracy, over a prescribed interval. (2) the characteristics of DD.

**Proof 1: Weierstrass approximation theorem [12], [13].** The expanded form of DD is a polynomial ($x_0 = 1$, where $x_0$ is an element of inputs). Based on Weierstrass approximation theorem, given any continuous function $f(x)$ on an interval $[a,b]$ and a tolerance $\epsilon > 0$, a polynomial $p_n(x)$ of sufficiently high degree $n$ exists, such that

$$|f(x) - p_n(x)| \leq \epsilon \quad \text{for all } x \in [a,b] \tag{1}$$

Therefore, DD can uniformly approximate any function that is merely continuous over a closed interval. Additionally, the function need not be analytic (nor differentiable) according to the above Weierstrass approximation theorem.

**Proof 2: Taylor's expansion and Error-backpropagation [14].** Consider first a function $f(x)$ of a single variable. Taylor's expansion for $f(x)$ about the point $a$ is

$$f(x) = f(a) + \frac{f'(a)}{1!}(x-a) + \frac{f''(a)}{2!}(x-a)^2 +$$
$$\frac{f'''(a)}{3!}(x-a)^3 + \cdots \frac{f^n(a)}{n!}(x-a)^n + R_n \quad (2)$$
$$= \sum_{n=0}^{N} \frac{f^{(n)}(a)}{n!}(x-a)^n + R_n$$

where $f^{(n)}(a)$ denotes the $n$-th derivative of $f$ evaluated at the point $a$. $R_n$ is the higher order infinitesimal of $(x-a)^n$. Suppose we have gotten $m$ points of $f(x)$. The Taylor's expansion at each point is as follows.

$$\begin{cases} f(x) = \sum_{n=0}^{N} \frac{f^{(n)}(a_1)}{n!}(x-a_1)^n + R_n \\ \vdots \\ f(x) = \sum_{n=0}^{N} \frac{f^{(n)}(a_m)}{n!}(x-a_m)^n + R_n \end{cases} \quad (3)$$

Then, $f(x)$ can be expressed as including Taylor's expansion with all sample points.

$$f(x) = \frac{1}{m}\Bigg(\sum_{n=0}^{N} \frac{f^{(n)}(a_1)}{n!}(x-a_1)^n +, \cdots$$
$$, + \sum_{n=0}^{N} \frac{f^{(n)}(a_m)}{n!}(x-a_m)^n \Bigg) + R_n$$
$$= \sum_{n=0}^{N}\Bigg(\frac{f^{(n)}(a_1)}{mn!}(x-a_1)^n +, \cdots$$
$$, + \frac{f^{(n)}(a_m)}{mn!}(x-a_m)^n\Bigg) + R_n \quad (4)$$
$$= \sum_{n=0}^{N}\Bigg(\begin{bmatrix} \frac{f^{(n)}(a_1)}{mn!} & \cdots & \frac{f^{(n)}(a_m)}{mn!} \end{bmatrix}$$
$$\begin{bmatrix} (x-a_1)^n \\ \cdots \\ (x-a_m)^n \end{bmatrix}\Bigg) + R_n$$

Eq 4 can also be generalized to functions of more than one variable.

$$f(x_1, \cdots, x_d) =$$
$$\sum_{n_1=0}^{N}\cdots\sum_{n_d=0}^{N}\Bigg(\begin{bmatrix} \frac{\left(\frac{\partial^{n_1+\cdots+n_d} f}{\partial x_1^{n_1}\cdots\partial x_d^{n_d}}\right)(a_{11},\cdots,a_{1d})}{m(n_1!\cdots n_d!)} \\ \vdots \\ \frac{\left(\frac{\partial^{n_1+\cdots+n_d} f}{\partial x_1^{n_1}\cdots\partial x_d^{n_d}}\right)(a_{m1},\cdots,a_{md})}{m(n_1!\cdots n_d!)} \end{bmatrix}^T$$
$$\begin{bmatrix} (x_1-a_{11})^{n_1}\cdots(x_d-a_{1d})^{n_d} \\ \vdots \\ (x_1-a_{m1})^{n_1}\cdots(x_d-a_{md})^{n_d} \end{bmatrix}\Bigg) + R_n \quad (5)$$

As is known from the above derivation, we found both fascinating characteristics.

(1) The coefficients of higher-order terms is suppressed by "$\frac{f^{(n)}(*)}{n!}$" in Taylor's expansion, which makes Taylor's expansion first use the lower order term for approximation and then use the higher-order term to compensate for the error, and the high-order term can be infinitesimal in infinite terms. It is well known that the higher-order terms of polynomials are unstable, and the high-order polynomial models are prone to overfitting in engineering applications. The characteristic of Taylor's expansion makes it avoid this problem.

(2) A function can be represented by a multi-point Taylor's expansion.

Additionally, in analogy to Fourier transform (trigonometric series, periodic components), the polynomial terms has physical meaning (the influence of inputs on the output). For example, polynomial terms contain $x_1$ (first-order independent component), $x_1^2$ (second-order independent component), and $x_1 x_2$ (second-order interactive component), etc., where $x_1$ and $x_2$ are inputs.

The special form of DD makes it inherit the above characteristics. The detail is as follows (see Fig. 1).

We set $Z^l = W^{l,l-1} A^{l-1}$. Then, DD module (***Eq 3 in the main paper***) can be expressed as follows.

$$\begin{cases} Z^l = W^{l,l-1} A^{l-1} \\ A^l = Z^l \circ X \end{cases} \quad (6)$$

The error backpropagation of the DD module based on the chain rule can be expressed as follows.

$$dA^{l-1}_{\text{prerious error}} = \frac{dA^l}{dA^{l-1}} dA^l_{\text{later error}}$$
$$= \frac{dZ}{dA^{l-1}} \cdot \left(\frac{dA^l}{dZ} \cdot dA^l_{\text{later error}}\right) \quad (7)$$
$$= (W^{l,l-1})^T \cdot \left(X \circ dA^l_{\text{later error}}\right)$$

where $dA^{l-1}_{\text{prerious error}}$ and $dA^l_{\text{later error}}$ denote the error of the front module and later module, respectively. It should be noted that $X$ denotes the inputs of Dendrite Net; thus, it is seen as a constant in the derivative of the single DD module (see Eq 7).

Then, $dZ^l = X \circ dA^l_{\text{later error}}$ is brought into ***Eq 7 in the main paper***. The weight adjustment can be expressed as follows.

$$\begin{cases} dW^{l,l-1} = \frac{1}{m} X \circ dA^l_{\text{later error}} \left(A^{l-1}\right)^T \\ W^{l,l-1(new)} = W^{l,l-1(old)} - \alpha dW^{l,l-1} \end{cases} \quad (8)$$

Since the expanded form of Dendrite Net is a polynomial, the derivative of Dendrite Net with respect to $X$ is composed of coefficients and Dendrite Net reduced by one order. For one DD module, the Dendrite Net reduced by one order happens to the inputs of the single modules. Let us look now at Eq 8, the "$dW^{l,l-1}$" contains "$A^{l-1}$" (the key of the derivative of Dendrite Net with respect to $X$), which is similar to "$f'(x)$" in Taylor's expansion. Additionally, becasue of $X \in [-1, 1]$, the suppression of "$X \circ dA^l_{\text{later error}}$" is similar

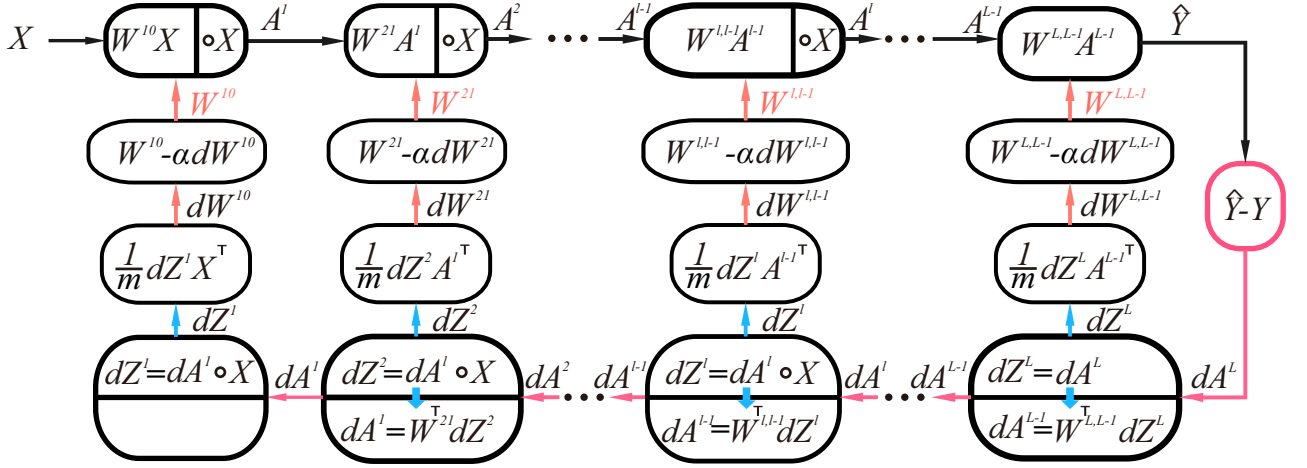

Fig. 1. Graphical illustration of learning rule.

to "$\frac{1}{n!}$". Therefore, DD suppresses the high-order terms, which is similar to Taylor's expansion.

Additionally, the derivation of computational complexity of Dendrite Net is briefly described as follows (see Fig. 1). The initial order (X) of the polynomial (the expanded form of Dendrite Net) is 1. The DD module ("$WA \circ X$") contains 2 matrix multiplication and can added 1 order to polynomial. The last linear module contains 1 matrix multiplication. Therefore, the computational complexity of Dendrite Net is $O(2n-1)$ where $n$ is the order of DD polynomial. As it is known, the computational complexity of polynomial is $O(n^2)$. Therefore, the computational complexity of PR is larger than DD module. Furthermore, the computational complexity of the nonlinear function is larger than multiplication. Therefore, the computational complexity of cell body module is larger than DD module.

Taken together, these theoretical derivations indicate DD's white-box attribute (the polynomial decompositions), better generalization capability (low-order approximation first and the suppression of high-order terms), and lower computational complexity ($O(2n-1)$).

### B. Notes and Tips

(1) The diagonal matrix can be selected as the initialized weight matrix, and the initialized weight should be as small as possible. (2) We recommend that the inputs and outputs are normalized to $[-1, 1]$. (3) When DD is upgraded, the weights can be initialized using low-order DD's trained weights. The recommended strategy is to gradually increase DD modules as the method indicated in Reference [15] (see Fig. 2). (4) We recommend adding a linear module in front of the DD module to increase or decrease the input data dimension to the dimension of the DD module (see *Fig. 11 in the main paper*). The dimension of the DD module is determined according to the complexity of the data. Generally, the simpler the data, the smaller the dimension. The number of DD modules is recommended to be gradually increased from 1 to find the optimal number.

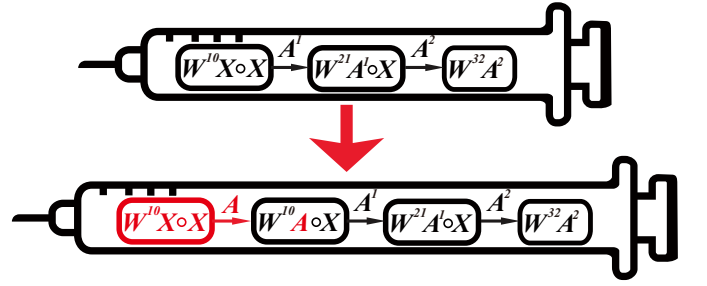

Fig. 2. Example of inheritance initialization. The red module is the added module. The weight matrix of the added module is initialized to a diagonal matrice with equal element values.

## III. SYSTEM IDENTIFICATION

### A. Identification comparison with Taylor's expansion

As an example, we select three-modules DD with two inputs and one output. To illustrate the process of formula simplification in software, we show it with labour.

$$\begin{aligned}
f(X) &= W^{32}(W^{21}((W^{10}X) \circ X) \circ X) \\
&= W^{32}\left(W^{21}\left(\begin{bmatrix} w_{11}^{10} & w_{12}^{10} \\ w_{21}^{10} & w_{22}^{10} \end{bmatrix}\begin{bmatrix} x_0 \\ x_1 \end{bmatrix} \circ \begin{bmatrix} x_0 \\ x_1 \end{bmatrix}\right) \circ X\right) \\
&= W^{32}\left(\begin{bmatrix} w_{11}^{21} & w_{12}^{21} \\ w_{21}^{21} & w_{22}^{21} \end{bmatrix}\begin{bmatrix} w_{11}^{10}x_0^2 + w_{12}^{10}x_0x_1 \\ w_{21}^{10}x_0x_1 + w_{22}^{10}x_1^2 \end{bmatrix} \circ \begin{bmatrix} x_0 \\ x_1 \end{bmatrix}\right) \\
&= \begin{bmatrix} w_{11}^{32} & w_{12}^{32} \end{bmatrix}\begin{bmatrix} w_{11}^{10}w_{11}^{21}x_0^3 + (w_{12}^{10}w_{11}^{21} + w_{21}^{10}w_{12}^{21})x_0^2x_1 \\ \quad + w_{22}^{10}w_{12}^{21}x_0x_1^2 \\ w_{11}^{10}w_{21}^{21}x_0^2x_1 + (w_{12}^{10}w_{21}^{21} + w_{21}^{10}w_{22}^{21})x_0x_1^2 \\ \quad + w_{22}^{10}w_{22}^{21}x_1^3 \end{bmatrix} \\
&= w_{11}^{10}w_{11}^{21}w_{11}^{32}x_0^3 + w_{22}^{10}w_{22}^{21}w_{12}^{32}x_1^3 \\
&\quad + (w_{12}^{10}w_{11}^{21}w_{11}^{32} + w_{21}^{10}w_{12}^{21}w_{11}^{32} + w_{11}^{10}w_{21}^{21}w_{12}^{32})x_0^2x_1 \\
&\quad + (w_{22}^{10}w_{12}^{21}w_{11}^{32} + w_{12}^{10}w_{21}^{21}w_{12}^{32} + w_{21}^{10}w_{22}^{21}w_{12}^{32})x_0x_1^2
\end{aligned} \quad (9)$$

where $x_0$ can be set as 1. Thus, the simplified DD consists of constant $c$, $x_1^3$, $x_1$, and $x_1^2$ items. The coefficients are in terms of DD's weights. For DD with more inputs and more modules, the result can be shown as a relation spectrum where the items and coefficients are the abscissa and ordinate, similar to Fourier spectrum.

## B. Identification for multiple-inputs system with noise

As an example, we select three-modules DD with three inputs and one output. To illustrate the process of formula simplification in software, we show it with labour.

$$
\begin{aligned}
f(X) &= W^{32}(W^{21}((W^{10}X) \circ X) \circ X) \\
&= W^{32}\left(W^{21}\left(\begin{bmatrix} w_{11}^{10} & w_{12}^{10} & w_{13}^{10} \\ w_{21}^{10} & w_{22}^{10} & w_{23}^{10} \\ w_{31}^{10} & w_{32}^{10} & w_{33}^{10} \end{bmatrix} \begin{bmatrix} x_0 \\ x_1 \\ x_2 \end{bmatrix} \circ \begin{bmatrix} x_0 \\ x_1 \\ x_2 \end{bmatrix}\right) \circ X\right) \\
&= W^{32}\left(\begin{bmatrix} w_{11}^{21} & w_{12}^{21} & w_{13}^{21} \\ w_{21}^{21} & w_{22}^{21} & w_{23}^{21} \\ w_{31}^{21} & w_{32}^{21} & w_{33}^{21} \end{bmatrix} \begin{bmatrix} w_{11}^{10}x_0^2 + w_{12}^{10}x_0x_1 \\ + w_{13}^{10}x_0x_2 \\ w_{21}^{10}x_0x_1 + w_{22}^{10}x_1^2 \\ + w_{23}^{10}x_1x_2 \\ w_{31}^{10}x_0x_2 + w_{32}^{10}x_1x_2 \\ + w_{33}^{10}x_2^2 \end{bmatrix} \circ \begin{bmatrix} x_0 \\ x_1 \\ x_2 \end{bmatrix}\right) \\
&= \begin{bmatrix} w_{11}^{32} & w_{12}^{32} & w_{13}^{32} \end{bmatrix} \begin{bmatrix} w_{11}^{10}w_{11}^{21}x_0^3 + (w_{12}^{10}w_{11}^{21} + w_{21}^{10}w_{12}^{21})x_0^2x_1 \\ + (w_{13}^{10}w_{11}^{21} + w_{31}^{10}w_{13}^{21})x_0^2x_2 + w_{22}^{10}w_{12}^{21}x_0x_1^2 \\ + (w_{23}^{10}w_{12}^{21} + w_{32}^{10}w_{13}^{21})x_0x_1x_2 + w_{33}^{10}w_{13}^{21}x_0x_2^2 \\[4pt]
w_{11}^{10}w_{21}^{21}x_0^2x_1 + (w_{12}^{10}w_{21}^{21} + w_{21}^{10}w_{22}^{21})x_0x_1^2 \\ + (w_{13}^{10}w_{21}^{21} + w_{31}^{10}w_{23}^{21})x_0x_1x_2 + w_{22}^{10}w_{22}^{21}x_1^3 \\ + (w_{23}^{10}w_{22}^{21} + w_{32}^{10}w_{23}^{21})x_1^2x_2 + w_{33}^{10}w_{23}^{21}x_1x_2^2 \\[4pt]
w_{11}^{10}w_{31}^{21}x_0^2x_2 + (w_{12}^{10}w_{31}^{21} + w_{21}^{10}w_{32}^{21})x_0x_1x_2 \\ + (w_{13}^{10}w_{31}^{21} + w_{31}^{10}w_{33}^{21})x_0x_2^2 + w_{22}^{10}w_{32}^{21}x_1^2x_2 \\ + (w_{23}^{10}w_{32}^{21} + w_{32}^{10}w_{33}^{21})x_1x_2^2 + w_{33}^{10}w_{33}^{21}x_2^3 \end{bmatrix} \\
&= w_{11}^{10}w_{11}^{21}w_{11}^{32}x_0^3 + w_{22}^{10}w_{22}^{21}w_{22}^{32}x_1^3 + w_{33}^{10}w_{33}^{21}w_{33}^{32}x_2^3 \\
&\quad + (w_{12}^{10}w_{11}^{21}w_{11}^{32} + w_{21}^{10}w_{12}^{21}w_{11}^{32} + w_{11}^{10}w_{21}^{21}w_{12}^{32})x_0^2x_1 \\
&\quad + (w_{13}^{10}w_{11}^{21}w_{11}^{32} + w_{31}^{10}w_{13}^{21}w_{11}^{32} + w_{11}^{10}w_{31}^{21}w_{13}^{32})x_0^2x_2 \\
&\quad + (w_{22}^{10}w_{12}^{21}w_{11}^{32} + w_{12}^{10}w_{21}^{21}w_{12}^{32} + w_{21}^{10}w_{22}^{21}w_{12}^{32})x_0x_1^2 \\
&\quad + (w_{33}^{10}w_{13}^{21}w_{11}^{32} + w_{13}^{10}w_{31}^{21}w_{13}^{32} + w_{31}^{10}w_{33}^{21}w_{13}^{32})x_0x_2^2 \\
&\quad + (w_{23}^{10}w_{22}^{21}w_{12}^{32} + w_{32}^{10}w_{23}^{21}w_{12}^{32} + w_{22}^{10}w_{32}^{21}w_{13}^{32})x_1^2x_2 \\
&\quad + (w_{33}^{10}w_{23}^{21}w_{12}^{32} + w_{23}^{10}w_{32}^{21}w_{13}^{32} + w_{32}^{10}w_{33}^{21}w_{13}^{32})x_1x_2^2 \\
&\quad + (w_{23}^{10}w_{12}^{21}w_{11}^{32} + w_{32}^{10}w_{13}^{21}w_{11}^{32} + w_{13}^{10}w_{21}^{21}w_{12}^{32} \\
&\qquad + w_{31}^{10}w_{23}^{21}w_{12}^{32} + w_{12}^{10}w_{31}^{21}w_{13}^{32} + w_{21}^{10}w_{32}^{21}w_{13}^{32})x_0x_1x_2
\end{aligned}
\tag{10}
$$

where $x_0$ can be set as 1. Thus, the simplified DD consists of constant $c$, $x_1^3$, $x_2^3$, $x_1$, $x_2$, $x_1^2$, $x_2^2$, $x_1^2x_2$, $x_1x_2^2$, and $x_1x_2$ items.

## IV. REGRESSION

### A. Datasets for regression

*1) Dataset 1: Yacht Hydrodynamics [16]:* Prediction of residuary resistance of sailing yachts at the initial design stage is of excellent value for evaluating the performance and estimating the required propulsive power. The dataset comprises 308 full-scale experiments that were performed at the Delft Ship Hydromechanics Laboratory. Attribute information includes six hull geometry coefficients, the Froude number, and the residuary resistance per unit weight of displacement. In this test, the model was used to predict the residuary resistance based on hull geometry coefficients and the Froude number.

*2) Dataset 2: Wine Quality [17]:* Wine certification is an essential link to assure the quality of the wine market. The quality evaluation is often part of the certification process and can improve winemaking and stratify wines such as premium brands. The dataset is considered, with a total of 4898 white samples. Attribute information includes eleven physicochemical data of wines and quality evaluation from wine taster. In this test, the model was used to predict white wine quality based on physicochemical data.

*3) Dataset 3: Behavior of the urban traffic [18]:* The prediction of the traffic behavior could help to make a decision about the routing process. The database was created with records of behavior of the urban traffic of the city of Sao Paulo in Brazil from December 14, 2009, to December 18, 2009 (From Monday to Friday), with a total of 135 samples. Attribute information includes seventeen behavior of urban traffic and slowness in traffic (%). In this test, the model was used to predict slowness in traffic based on the behavior of urban traffic.

*4) Dataset 4: QSAR fish toxicity [19]:* To prove that products are safe for both human health and the environment, REACH requires the evaluation of the short-term toxic effects on fish for substances imported or manufactured in quantities greater than 10 tonnes per year. The dataset is acute aquatic toxicity towards the fish Pimephales promelas (fathead minnow) on a set of 908 chemicals. In this test, the model was used to predict LC50 data that is the concentration that causes death in 50% of test fish over a test duration of 96 hours.

*5) Dataset 5: Auto MPG [16]:* This dataset is a slightly modified version of the dataset provided in the StatLib library. The data concerns city-cycle fuel consumption in miles per gallon, to be predicted in terms of three multivalued discrete and five continuous attributes [20]. In this paper, we removed a few samples that include unknown values and remained 392 samples. The model was used to predict "mpg".

*6) Dataset 6: Stock portfolio performance [21]:* The return of investment portfolios can be increased by adopting the appropriate factors in stock. The database includes six weights of stock-picking concepts and six performances of portfolios with 315 samples from US stock market historical data. In this test, the model was used to predict the relative winning rate based on stock-picking concepts' weights.

*7) Dataset 7: Real estate valuation [22]:* The market historical data set of real estate valuation is collected from Sindian Dist, New Taipei City, Taiwan, with 414 samples. Attribute information includes house price of the unit area and six attribute variables of house, such as the transaction date, the house age, and the nearest MRT station's distance. In this test, the model was used to predict the unit area's house price based on the attribute variables of houses.

*8) Dataset 8: Daily Demand Orders [23]:* The dataset was collected a real database of a Brazilian logistics company for 60 days. The data has twelve predictive attributes and a target that is the total of orders for daily treatment.

*9) Dataset 9: Concrete Compressive Strength [24]:* Concrete is the most important material in civil engineering. The dataset includes eight concrete attributes and concrete compressive strength, with 1030 samples. In this test, the model was applied to predict the concrete compressive strength.

### B. Experiments and results

In this paper, we aim to explore the impact of as many data factors as possible on generalization capability, such as the number of training samples. Thus, we did not select cross-validation test methods used in previous literature (i.e., 5-fold cross-validation, 10-fold cross-validation or jackknife cross-validation test) and took the following exhaustive approach

under the independent dataset test (*See supplementary materials* **Algorithm 1: Testing the performance under different numbers of training samples.** ) [25].

---

**Algorithm 1:** DD vs. Cell body Net vs. PR vs. SVM

**Input:** Dataset
**Output:** MSE

1 **for** *i = 1 to 100* **do**
2     **for** *n = t to end with c interval* **do**
3        Randomly choose *n* samples from Dataset as the training samples;
4        Choose residual data as the testing samples;
5        Train DD;
6        Train Cell body Net;
7        Train PR;
8        Train SVM;
9        Test DD;
10       Test Cell body Net;
11       Test PR;
12       Test SVM;

---

## V. CLASSIFICATION

### A. Datasets for classification

*1) Dataset 1: MNIST [26]:* The MNIST database of handwritten digits has a training set of 60,000 examples and a test set of 10,000 examples. It is a subset of a larger set available from NIST. The digits have been size-normalized and centered in a fixed-size image.

*2) Dataset 2: FASHION-MNIST [27]:* FASHION-MNIST is a dataset of Zalando's paper images—consisting of a training set of 60,000 examples and a test set of 10,000 examples. Each example is a 28x28 grayscale image, associated with a label from 10 classes.

## REFERENCES

[1] P. Geladi and B. R. Kowalski, "Partial least-squares regression: a tutorial," *Analytica chimica acta*, vol. 185, pp. 1–17, 1986.
[2] P. Peduzzi, J. Concato, E. Kemper, T. R. Holford, and A. R. Feinstein, "A simulation study of the number of events per variable in logistic regression analysis," *Journal of clinical epidemiology*, vol. 49, no. 12, pp. 1373–1379, 1996.
[3] P. N. Belhumeur, J. P. Hespanha, and D. J. Kriegman, "Eigenfaces vs. fisherfaces: Recognition using class specific linear projection," *IEEE Transactions on pattern analysis and machine intelligence*, vol. 19, no. 7, pp. 711–720, 1997.
[4] J. R. Quinlan, "Induction of decision trees," *Machine learning*, vol. 1, no. 1, pp. 81–106, 1986.
[5] Q. Wang, G. M. Garrity, J. M. Tiedje, and J. R. Cole, "Naive bayesian classifier for rapid assignment of rrna sequences into the new bacterial taxonomy," *Applied and environmental microbiology*, vol. 73, no. 16, pp. 5261–5267, 2007.
[6] C.-C. Chang and C.-J. Lin, "Libsvm: A library for support vector machines," *ACM transactions on intelligent systems and technology (TIST)*, vol. 2, no. 3, pp. 1–27, 2011.
[7] J. Friedman, T. Hastie, R. Tibshirani *et al.*, "Additive logistic regression: a statistical view of boosting (with discussion and a rejoinder by the authors)," *The annals of statistics*, vol. 28, no. 2, pp. 337–407, 2000.
[8] J. A. Hartigan and M. A. Wong, "Algorithm as 136: A k-means clustering algorithm," *Journal of the royal statistical society. series c (applied statistics)*, vol. 28, no. 1, pp. 100–108, 1979.
[9] L. Breiman, "Random forests," *MACHINE LEARNING*, vol. 45, no. 1, pp. 5–32, OCT 2001.
[10] K. Hornik, M. Stinchcombe, H. White *et al.*, "Multilayer feedforward networks are universal approximators." *Neural networks*, vol. 2, no. 5, pp. 359–366, 1989.
[11] G. Bao, Y. Zhang, and Z. Zeng, "Memory analysis for memristors and memristive recurrent neural networks," *IEEE/CAA Journal of Automatica Sinica*, vol. 7, no. 1, pp. 96–105, 2019.
[12] R. T. Farouki, "The bernstein polynomial basis: A centennial retrospective," *Computer Aided Geometric Design*, vol. 29, no. 6, pp. 379–419, 2012.
[13] M. H. Stone, "The generalized weierstrass approximation theorem," *Mathematics Magazine*, vol. 21, no. 5, pp. 237–254, 1948.
[14] M. Abramowitz, I. A. Stegun *et al.*, *Handbook of mathematical functions: with formulas, graphs, and mathematical tables*. National bureau of standards Washington, DC, 1972, vol. 55.
[15] G. E. Hinton, S. Osindero, and Y.-W. Teh, "A fast learning algorithm for deep belief nets," *Neural computation*, vol. 18, no. 7, pp. 1527–1554, 2006.
[16] D. Dua and C. Graff, "UCI machine learning repository," 2017. [Online]. Available: http://archive.ics.uci.edu/ml
[17] P. Cortez, A. Cerdeira, F. Almeida, T. Matos, and J. Reis, "Modeling wine preferences by data mining from physicochemical properties," *Decision Support Systems*, vol. 47, no. 4, pp. 547 – 553, 2009, smart Business Networks: Concepts and Empirical Evidence. [Online]. Available: http://www.sciencedirect.com/science/article/pii/S0167923609001377
[18] C. Affonso, R. J. Sassi, and R. P. Ferreira, "Traffic flow breakdown prediction using feature reduction through rough-neuro fuzzy networks," in *The 2011 International Joint Conference on Neural Networks*. IEEE, 2011, pp. 1943–1947.
[19] M. Cassotti, D. Ballabio, R. Todeschini, and V. Consonni, "A similarity-based qsar model for predicting acute toxicity towards the fathead minnow (pimephales promelas)," *SAR and QSAR in Environmental Research*, vol. 26, no. 3, pp. 217–243, 2015.
[20] J. R. Quinlan, "Combining instance-based and model-based learning," in *Proceedings of the tenth international conference on machine learning*, 1993, pp. 236–243.
[21] Y.-C. Liu and I.-C. Yeh, "Using mixture design and neural networks to build stock selection decision support systems," *Neural Computing and Applications*, vol. 28, no. 3, pp. 521–535, 2017.
[22] I.-C. Yeh and T.-K. Hsu, "Building real estate valuation models with comparative approach through case-based reasoning," *Applied Soft Computing*, vol. 65, pp. 260–271, 2018.
[23] R. P. Ferreira, A. Martiniano, A. Ferreira, and R. J. Sassi, "Study on daily demand forecasting orders using artificial neural network," *IEEE Latin America Transactions*, vol. 14, no. 3, pp. 1519–1525, 2016.
[24] I.-C. Yeh, "Modeling of strength of high-performance concrete using artificial neural networks," *Cement and Concrete research*, vol. 28, no. 12, pp. 1797–1808, 1998.
[25] J. D. Rodriguez, A. Perez, and J. A. Lozano, "Sensitivity analysis of k-fold cross validation in prediction error estimation," *IEEE transactions on pattern analysis and machine intelligence*, vol. 32, no. 3, pp. 569–575, 2009.
[26] Y. LeCun, L. Bottou, Y. Bengio, and P. Haffner, "Gradient-based learning applied to document recognition," *Proceedings of the IEEE*, vol. 86, no. 11, pp. 2278–2324, 1998.
[27] H. Xiao, K. Rasul, and R. Vollgraf, "Fashion-mnist: a novel image dataset for benchmarking machine learning algorithms," *arXiv preprint arXiv:1708.07747*, 2017.



\end{document}